\documentclass[10pt,twocolumn,letterpaper]{article}

\usepackage[pagenumbers]{arxiv_twocolumn} 

\usepackage{fontawesome5}

\definecolor{cvprblue}{rgb}{0.21,0.49,0.74}
\usepackage[pagebackref,breaklinks,colorlinks,allcolors=cvprblue]{hyperref}

\def\paperID{*****} 
\def\confName{CVPR}
\def\confYear{2026}

\definecolor{newgray}{gray}{0.4}

\newcommand{\actionicon}{\raisebox{-0.34em}{\includegraphics[height=1.3em]{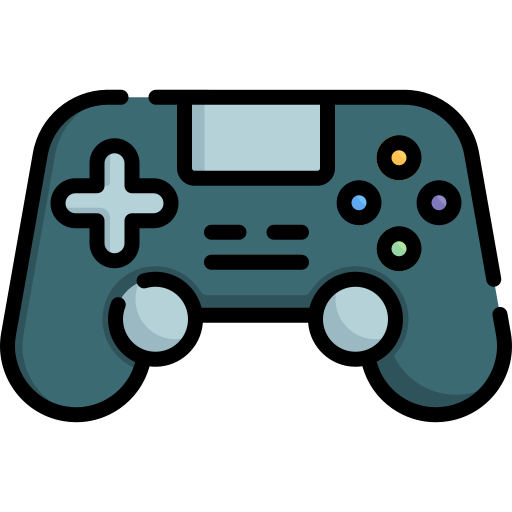}}}
\newcommand{\cameraicon}{\raisebox{-0.24em}{\includegraphics[height=1.2em]{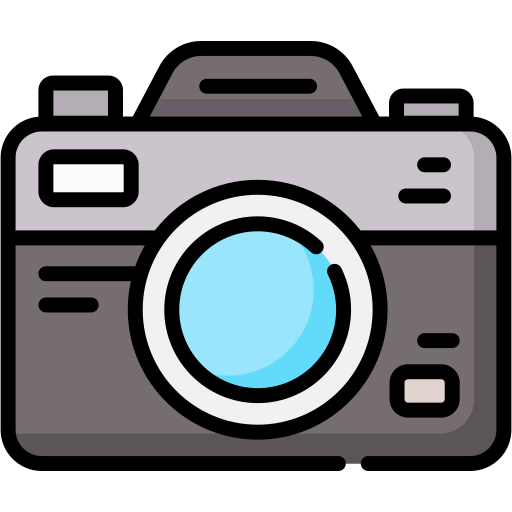}}}

\newcommand{\tongyiicon}{\raisebox{-0.24em}{\includegraphics[height=1.4em]{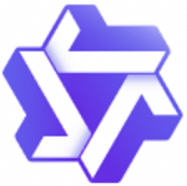}}}

\definecolor{headingpurple}{HTML}{7030A0}

\hypersetup{
    colorlinks=true,
    linkcolor=headingpurple,
    citecolor=headingpurple!50,
    filecolor=headingpurple,      
    urlcolor=headingpurple,
    }

\patchcmd{\cvprsection}
  {\raggedright}
  {\raggedright\color{headingpurple}}
  {}
  {\PackageWarning{heading-colors}{Failed to patch section}}

\patchcmd{\paragraph}
  {\bfseries}
  {\bfseries\color{headingpurple}}
  {}
  {\PackageWarning{heading-colors}{Failed to patch paragraph}}

\patchcmd{\cvprsubsection}
  {\raggedright}
  {\raggedright\color{headingpurple}}
  {}
  {\PackageWarning{heading-colors}{Failed to patch subsection}}

\patchcmd{\abstract}
  {\centerline{\large\bf Abstract}}
  {\centerline{\large\bf\color{headingpurple} Abstract}}
  {}
  {\PackageWarning{heading-colors}{Failed to patch abstract}}

\title{World-in-Context: Interactive Video World Modeling with In-Context Conditioning}

\author{
\vspace{-2em}\\
\fontsize{10.4pt}{9.84pt}\selectfont
\tongyiicon~Alibaba Token Hub, Alibaba Group\\
\\
\fontsize{9.8pt}{9pt}\selectfont
{\faGlobe ~}{\href{https://haroldchen19.github.io/WiC-site}{https://haroldchen19.github.io/WiC-site}}~~~
{\faGithub ~}{\href{https://github.com/aigc3d/World-in-Context}{https://github.com/aigc3d/World-in-Context}}
}

\begin{document}
\twocolumn[{
\renewcommand\twocolumn[1][]{#1}
\maketitle
\begin{center}
    \vspace{-14pt}
   \includegraphics[width=\linewidth]{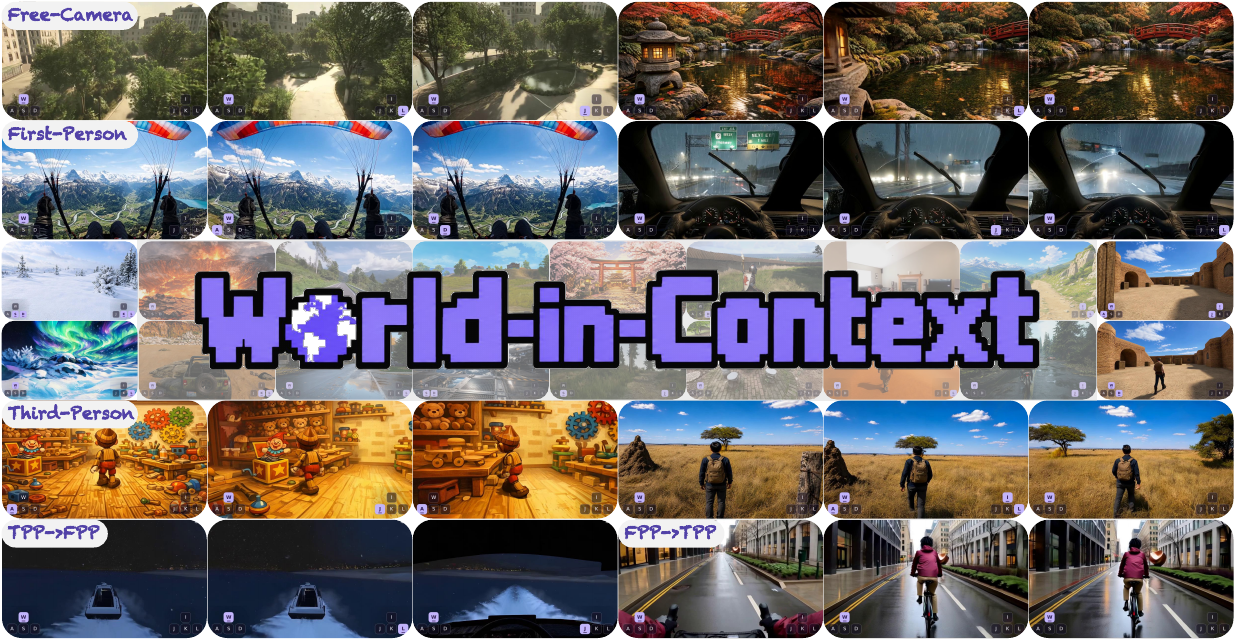}
   \vspace{-1.8em}
    \captionof{figure}{\label{fig:teaser}
    \ourmethod supports versatile interactive video generation across \textsc{Free}, \textsc{Fpp}, and \textsc{Tpp} regimes, supporting dynamic in-video view switching under joint action and camera control.}
\end{center}
}]

\begin{abstract}
Interactive video world models require coordinated control over viewing perspectives, character actions, and camera motion.
Existing methods typically leave view modes implicit in text captions, conflate action with camera motion, and inject controls through fragmented architectural pathways that capture interdependencies only indirectly.
In this work, we present \ourmethod, an interactive world modeling framework powered by In-Context Conditioning (ICC). 
Rather than relying on specialized conditioning pathways, \ourmethod unifies view, action, and camera controls into explicit tokens contextualized directly within a shared self-attention space. 
Our formulation reveals that (\textbf{\textit{i}}) explicit view tokens disentangle conflicting dynamics and unlock emergent view switching from single-view data, while (\textbf{\textit{ii}}) action and camera inputs play complementary roles in governing local responsiveness and global stability. 
To preserve temporal continuity across execution intervals, we further introduce Interaction-Transport RoPE (IT-RoPE), parameterizing temporal attention with accumulated control transitions. 
On WBench, \ourmethod-5B achieves navigation performance on par with 22B EchoWM ($81.5$ \textit{vs.} $81.6$) and attains a $93.8\%$ perspective-switching success rate, substantially outperforming existing 5B baselines ($18.8\%$). 
Controlled analyses further demonstrate compounding advantages of \ourmethod under richer conditioning mixtures, favorable data efficiency, and natural applicability to causal rollouts.

\end{abstract}


\vspace{-1.4em}
\section{Introduction}
\label{sec:intro}

Video generation has rapidly advanced from passive video generation \citep{wan2025wan, hacohen2024ltx, kong2024hunyuanvideo} toward interactive world models \citep{team2026advancing, gao2026infinite, team2026alayaworld, wang2026matrix}, empowering users to actively steer dynamic visual environments. In this paradigm, the model acts as an interactive simulator where users provide control inputs to navigate and explore the environment, while the model dynamically predicts the evolving visual observations in response. At its core, effectively simulating such an interactive world hinges on two fundamental dimensions \citep{liu2026towards, wang2026mechanistic}: \ding{192}~\textit{how the world is viewed} across distinct perspectives, and \ding{193}~\textit{how user inputs concurrently govern character dynamics and camera viewports}. Together, these dimensions define how an agent actively perceives and intervenes in an unfolding generative world.

Despite recent progress \citep{team2026advancing, zhang2026echowm, zhu2026sana, team2026dreamx, huang2026solarwm}, current interactive pipelines often overlook the systematic formulation of these two essential dimensions: \ding{182}~\textbf{view-mode ambiguity}: first- (\textsc{Fpp}) and third-person (\textsc{Tpp}) perspectives increasingly appear in interactive data, yet view mode is often specified only implicitly through text captions \cite{gao2026infinite, team2026alayaworld, team2026dreamx, zhang2026echowm}, while free-camera (\textsc{Free}) roaming is frequently conflated with \textsc{Fpp} control despite their different motion semantics; and \ding{183}~\textbf{action-camera entanglement}: most pipelines rely primarily on continuous camera trajectories alone for control \cite{nam2026worldcam, team2026dreamx, zhang2026echowm, huang2026disco}, overlooking the distinct role of discrete actions, especially in third-person interaction where character and camera motion form separate degrees of freedom.

\begin{figure*}[!t]
\centering
\vspace{-1em}
\includegraphics[width=\linewidth]{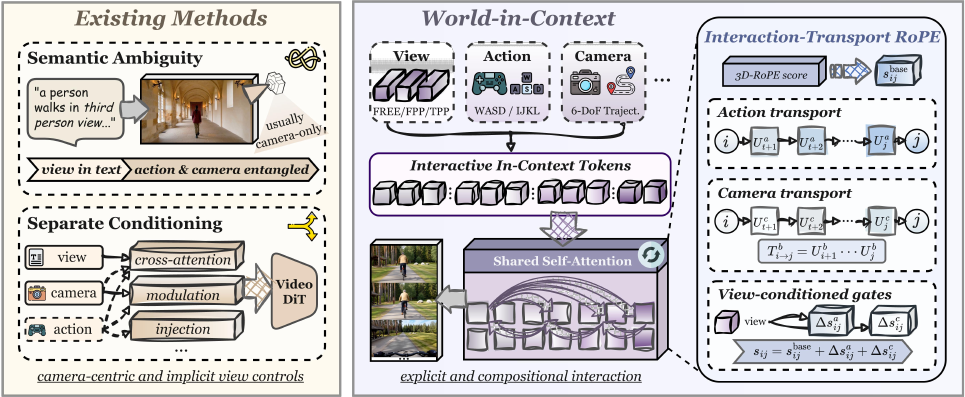}
\vspace{-1.8em}
\caption{(\textbf{\textit{Left}}) Prior methods use implicit view descriptions and separate control pathways. (\textbf{\textit{Right}}) \ourmethod unifies controls through ICC and models their temporal evolution with IT-RoPE.}
\label{fig:pipleine}
\vspace{-1em}
\end{figure*}

A naive remedy would be simply attaching auxiliary condition channels for each missing signal. However, existing paradigms \citep{team2026advancing, gao2026infinite, jiang2026abot, zhao2026minwm, qian2026matrix, wang2026matrix} typically accommodate interactive controls through separate modules (\textit{e.g.}, AdaLN modulation), which tends to isolate different control signals and leaves their cross-dependencies to be captured only indirectly, as shown in \Cref{fig:pipleine} (\textit{Left}). As interaction becomes richer and more compositional, coordinating these modular branches grows increasingly intricate, especially when the interpretation of one signal inherently depends on concurrent inputs and the active viewport.

In this work, we propose \ourmethod, which unifies multi-dimensional interaction through an \textit{In-Context Conditioning} (ICC) paradigm. Rather than routing controls into isolated architectural branches, \ourmethod casts diverse interactive signals into token sequences and prepends them directly into the shared self-attention space alongside visual latents. Within this unified interface, we revisit the aforementioned limitations along two fronts:
\ding{168}~\textbf{Explicit view-mode modeling}: Rather than leaving perspectives implicit in text captions \citep{gao2026infinite, team2026advancing, team2026alayaworld, zhu2026sana}, we explicitly incorporate discrete view tokens (\textsc{Free}, \textsc{Fpp}, and \textsc{Tpp}) into the context. Joint training across these regimes distinguishes their motion semantics, while enabling \textit{in-video view switching to emerge naturally}.
\ding{171}~\textbf{Joint action-camera modeling}:  We jointly incorporate discrete actions and continuous camera trajectories into the context.
Crucially, beyond their necessary separation in \textsc{Tpp} interaction, we further study the less-examined \textsc{Fpp} and \textsc{Free} regimes, where camera trajectories often stand in for both forms of control \citep{nam2026worldcam, huang2026solarwm, team2026dreamx}. Our analysis reveals complementary roles: action governs \textit{local command fidelity}, whereas camera anchors \textit{global trajectory stability}.

Built upon ICC, \ourmethod natively captures cross-condition dependencies within a unified attention space.
While in-context tokens specify what controls are present at individual frames, naive token conditioning can still leave the dynamic evolution between distant moments to be modeled implicitly.
To this end, we further introduce \textbf{IT-RoPE} (Interaction-Transport RoPE), a lightweight positional encoding that incorporates pairwise control transitions directly into temporal attention.
By jointly modeling control conditions and their temporal evolution, \ourmethod supports coherent interaction across viewing regimes and makes effective use of interactive supervision, with increasing benefits as controls become richer. More broadly, it reframes multi-condition interaction as context modeling, offering a unified interface for interactive world modeling.

To summarize, this work contributes threefold: 
\begin{itemize}[leftmargin=*]
    \item[\ding{226}] We introduce In-Context Conditioning (ICC) for interactive world modeling and present \ourmethod, which represents view, action, and camera controls as tokens within a shared attention space, providing a unified interface for diverse interaction signals.
    \item[\ding{226}] We investigate the often-overlooked roles of viewing regimes and joint action-camera control, revealing distinct view-dependent dynamics, emergent view switching, and complementary action-camera roles in command fidelity and trajectory stability. 
    \item[\ding{226}]
    Extensive experiments on WBench demonstrate that \ourmethod-5B matches the 22B EchoWM in navigation ($81.5$ \textit{vs.} $81.6$) and substantially outperforms existing 5B baselines in perspective switching ($93.8\%$ \textit{vs.} $18.8\%$). Controlled studies further confirm compounding gains of \ourmethod under richer controls, favorable sample efficiency, and causal applicability.
\end{itemize}

\vspace{-0.4em}
\section{Related Work}
\vspace{-0.4em}

\paragraph{Interactive Video World Models.} Interactive world models extend video generation toward closed-loop interaction, where future observations evolve according to user-steered conditions \citep{parkerholder2024genie2, decart2024oasis, chen2026past, sun2025worldplay, wang2026matrix}. Following recent formulations \citep{liu2026towards, wang2026mechanistic}, existing models mainly navigate two core dimensions:
(\textbf{\textit{i}})~\textbf{How the world is viewed.}
Interactive environments span \textsc{Free}, \textsc{Fpp}, and \textsc{Tpp} regimes. Prior works \citep{gao2026infinite, team2026alayaworld, team2026dreamx, zhang2026echowm} typically subsume them into a shared camera formulation, leaving perspective implicit in text captions and often conflating \textsc{Free} roaming with \textsc{Fpp} motion. We instead explicitly condition on discrete view modes, distinguishing their physical motion semantics and naturally enabling dynamic view switching.
(\textbf{\textit{ii}})~\textbf{How user inputs affect its evolution.}
Existing systems rely almost exclusively on continuous camera trajectories for navigation \citep{nam2026worldcam, zhu2026sana, team2026dreamx, gao2026infinite, zhang2026echowm, huang2026disco} for all view modes, which inevitably entangles character and camera dynamics in \textsc{Tpp}. 
To incorporate richer signals, recent approaches either formulate a unified action interface across different regimes \citep{jiang2026abot}, or inject actions and camera poses through distinct architectural pathways \citep{sun2025worldplay}. 
Differing from these paradigms, we explicitly maintain view modes and represent character actuation alongside camera motion as jointly contextualized controls within the same attention space, effectively treating them as coexisting yet independent degrees of freedom.

\vspace{-1em}
\paragraph{In-Context Conditioning.}
In-context learning, originally developed for language models \cite{dong2024survey}, has increasingly been explored in visual generation. Prompt Diffusion \cite{wang2023context} first enables diffusion models to infer visual tasks from contextual examples, followed by works that exploit shared attention contexts for task-agnostic or unified image generation \cite{huang2024group, huang2024context, xiao2025omnigen, tan2025ominicontrol, wu2025omnigen2}. This paradigm has recently extended to video through spatial or temporal in-context generation \cite{fei2024video, kim2025temporal}, unified multimodal conditioning with full attention \cite{ju2025fulldit, he2025fulldit2}, and reference videos as semantic prompts \cite{bian2026video}. IC-World \cite{wu2026ic} further explores in-context generation for shared world modeling, focusing on consistency across multiple observations of the same world. In this work, we extend in-context conditioning to interactive video world modeling.

\section{Preliminary: Problem Formulation}
\vspace{-0.4em}

Let $\mathbf{x} \in \mathbb{R}^{T \times H \times W \times 3}$ denote a video clip of $T$ frames, and $\mathbf{y}$ the initial contextual specification (\eg, a text prompt or a starting frame). In interactive video world modeling, the model generates future visual observations conditioned on temporally aligned user interaction sequences $\mathbf{u} = \{\mathbf{u}_t\}_{t=1}^T$, parameterizing the conditional distribution:
\setlength\abovedisplayskip{3.4pt}
\setlength\belowdisplayskip{3.4pt}
\begin{equation}
\label{eq:world_model_formulation}
p_\theta(\mathbf{x} \mid \mathbf{y}, \mathbf{u}).
\end{equation}
Modern video diffusion and flow-matching backbones typically instantiate this conditional process within a latent space. Specifically, let $\mathbf{z}_\tau$ denote the latent video representations at diffusion timestep $\tau \in [0, 1]$. Existing models typically inject interactive conditions $\mathbf{u}$ via modular, condition-specific pathways:
\begin{equation}
\label{eq:side_channel_cond}
\hat{\mathbf{v}}_\tau = \mathcal{F}_\theta\big(\mathbf{z}_\tau, f_{\mathrm{cond}}(\mathbf{u}); \mathbf{y}, \tau\big),
\end{equation}
where $f_{\mathrm{cond}}(\cdot)$ commonly instantiates modulation for camera trajectories \citep{gao2026infinite, hyworld2025}, additive token injection for discrete actions \citep{jiang2026abot, chen2026zing}, cross-attention for text-form controls \citep{mao2026yume1}, or geometry-aware positional attention and warping for camera control \citep{zhang2026echowm, team2026dreamx, yin2026alaya}. While effective for isolated signals, such architectural separation isolates diverse controls into disparate parameter spaces, hindering the model from capturing cross-condition dependencies directly. In this work, instead of introducing dedicated condition branches, we reframe multi-modal interactive conditioning as a unified in-context sequence modeling problem:
\begin{equation}
\label{eq:icc_formulation}
\hat{\mathbf{v}}_\tau = \mathcal{F}_\theta\big([\mathbf{c} \mathbin{\Vert} \mathbf{z}_\tau]; \mathbf{y}, \tau\big),
\end{equation}
where diverse controls are recast as unified condition tokens $\mathbf{c}$ and natively contextualized with visual latents within the shared attention manifold.

\begin{figure*}[!t]
\centering
\includegraphics[width=\linewidth]{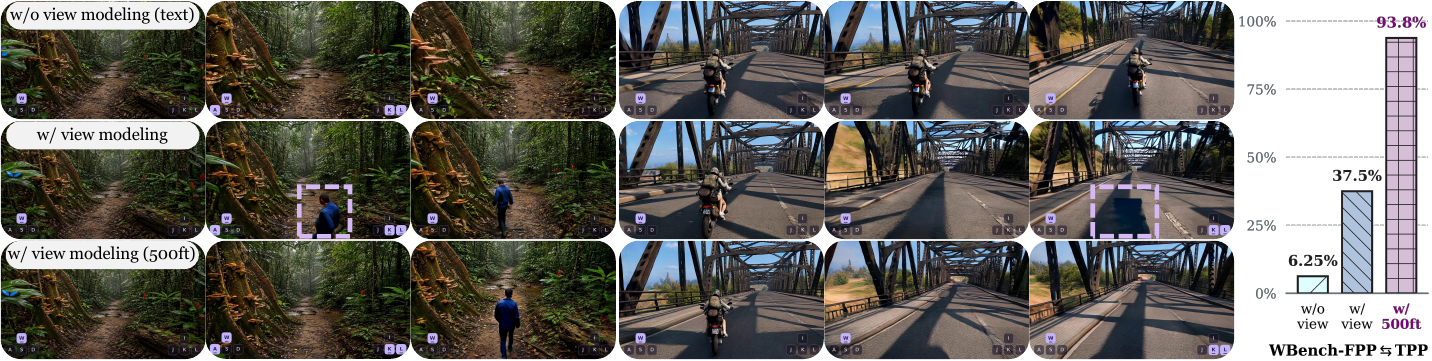}
\vspace{-2em}
\caption{\textbf{Emergent view switching.} (\textbf{\textit{Left}}) Explicit view modeling enables switching with residual artifacts; fine-tuning on 500 switching clips (``500ft") improves transitions. (\textbf{\textit{Right}}) Success rates on WBench \textsc{Fpp}$\leftrightarrow$\textsc{Tpp} set \citep{ying2026wbench}.}
\label{fig:42_b}
\vspace{-1em}
\end{figure*}

\section{World-in-Context}
\label{sec:method}

\subsection{Overview}
To avoid the scalability bottleneck of fragmented side channels (\eg, AdaLN or separate cross-attention layers), \ourmethod casts diverse interactive signals directly into sequence tokens within a unified attention space, as shown in \Cref{fig:pipleine} (\textit{Right}). The world modeling space is organized around two complementary pillars: (\textbf{\textit{i}})~\textit{how the world is viewed} across perspectives, and (\textbf{\textit{ii}})~\textit{how user inputs steer dynamics} through concurrent character action and camera viewport movement.

\vspace{-1em}
\paragraph{Unified In-Context Conditioning.}
Instead of introducing specialized parameter pathways for each control modality, all interactive signals are projected into a unified sequence of condition tokens $\mathbf{c} \in \mathbb{R}^{L_c \times D}$, matching the backbone hidden dimension $D$:
\begin{equation}
    \mathbf{c} = \big[ \mathbf{c}^{\mathrm{view}} \,;\, \mathbf{c}^{\mathrm{action}} \,;\, \mathbf{c}^{\mathrm{camera}} \big],
    \label{eq:unified_condition}
\end{equation}
where $[\,\cdot\,;\,\cdot\,]$ denotes concatenation along the sequence dimension. These condition tokens are prepended directly to the visual latents $\mathbf{z}_\tau$ at diffusion timestep $\tau \in [0, 1]$. The flow-matching generative backbone then predicts the velocity field $\hat{\mathbf{v}}_\tau$ over this shared sequence:
\begin{equation}
    \hat{\mathbf{v}}_\tau = \mathcal{F}_\theta\big( [\mathbf{c} \,;\, \mathbf{z}_\tau] \,;\, \mathbf{y}, \tau \big).
    \label{eq:in_context_backbone}
\end{equation}
Ingesting all controls into the shared self-attention allows cross-condition dependencies to be captured natively without ad-hoc modules, reframing interactive expansion into an in-context modeling problem.

\vspace{-1em}
\paragraph{Decomposing Interactive Context.}
The condition sequence $\mathbf{c}$ of \ourmethod explicitly decouples the interactive degrees of freedom:
\begin{itemize}[leftmargin=1.4em,itemsep=1pt,topsep=1pt]
    \item[\ding{111}] \textbf{View Mode} ($\mathbf{c}^{\mathrm{view}}$): Disentangles \textsc{Free}, \textsc{Fpp}, and \textsc{Tpp} regimes via learned categorical tokens, mitigating view ambiguity and naturally supporting dynamic switching.
    \item[\ding{111}] \textbf{Action} ($\mathbf{c}^{\mathrm{action}}$): Encodes hybrid keyboard-action inputs to govern local character actuation, providing responsive command dynamics.
    \item[\ding{111}] \textbf{Camera} ($\mathbf{c}^{\mathrm{camera}}$): Embeds relative 6-DoF camera poses and Pl\"ucker coordinates to ensure continuous trajectory tracking and viewpoint stability.
\end{itemize}
To complement in-context conditioning, we further incorporates geometric transport into the temporal attention mechanism to preserve continuous physical transitions across frames. The subsequent sections elaborate on the explicit view-mode formulation (\S\ref{sec:view_mode}), examine joint action-camera decoupling (\S\ref{sec:action_camera}), and introduce IT-RoPE for parameterized temporal attention (\S\ref{sec:rope}).

\subsection{Explicit View-Mode Modeling}
\label{sec:view_mode}

Most existing models \citep{team2026advancing, nam2026worldcam, team2026alayaworld, zhao2026minwm, gao2026infinite} convey viewing regimes either through implicit text descriptions $\mathbf{y}$ or directly via camera poses. 
However, identical geometric trajectories inherently entail different motion semantics across regimes, for example, distinguishing unconstrained \textsc{Free} navigation from embodied \textsc{Fpp} motion. Leaving such distinctions to implicit text conditioning makes it ambiguous to resolve regime-dependent motion characteristics from identical geometric inputs.

We therefore restrict $\mathbf y$ strictly to world-state specifications and introduce explicit, time-aligned view tokens $\mathbf c^{\mathrm{view}} = \{\mathbf c_t^{\mathrm{view}}\}_{t=0}^{T_\mathbf{z}-1}$ into the shared context:
\begin{equation}
\mathbf c_t^{\mathrm{view}} = \mathbf e_{\mathrm{view}}(v_t) + \mathbf e_{\mathrm{type}}(\mathrm{view}),~~v_t \in \{\textsc{Free}, \textsc{Fpp}, \textsc{Tpp}\}.
\end{equation}
By attending directly to visual latents, $\mathbf c_t^{\mathrm{view}}$ acts as a contextual switch that modulates the shared attention space toward regime-appropriate physical motion fields. 

To validate the efficacy of explicit view conditioning, we compare our formulation against a baseline that specifies view modes through text alone (``w/o view modeling") 
\Cref{fig:42_a} evaluates trajectory oscillations under \textsc{Fpp} and \textsc{Free} conditioning across $100$ scenes, while \Cref{fig:42_b} presents qualitative transitions alongside quantitative evaluations on the WBench \textsc{Fpp}$\leftrightarrow$\textsc{Tpp} suite. 
Detailed setups are provided in \Cref{app_sec:method_setup}; key observations are summarized:

\begin{figure}[!t]
\centering
\includegraphics[width=\linewidth]{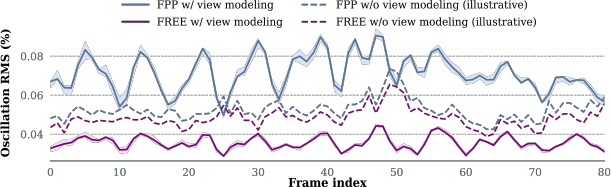}
\vspace{-2em}
\caption{\textbf{View-dependent motion dynamics.} Oscillation RMS across 100 scenes. Explicit view modeling distinguishes \textsc{Fpp} oscillations from smoother \textsc{Free} motion.}
\label{fig:42_a}
\vspace{-1.6em}
\end{figure}

\noindent\textbf{Obs.\ding{182} Explicit view modeling distinguishes regime-dependent motion dynamics.}\quad As shown in \Cref{fig:42_a}, text-based conditioning yields closely overlapping oscillation profiles for \textsc{Fpp} and \textsc{Free}, conflating their respective physical semantics. 
In contrast, explicit view modeling encourages a distinct separation: \textsc{Fpp} exhibits pronounced periodic oscillations characteristic of embodied locomotion, while \textsc{Free} preserves smoother camera trajectories. 
This indicates that explicit view tokens effectively mitigate motion ambiguity across viewing regimes.

\vspace{0.4em}
\noindent\textbf{Obs.\ding{183} In-video view switching emerges naturally from explicit view tokens.}\quad Because view tokens act as modular contextual switches within the shared attention space, training exclusively on single-view clips already gives rise to perspective transitions simply by varying the view token sequence during rollout, as shown in \Cref{fig:42_b}. 
On the WBench \textsc{Fpp}$\leftrightarrow$\textsc{Tpp} set, our base model without switching supervision achieves a $37.5\%$ success rate, substantially outperforming the text-based baseline ($6.25\%$). 
Incorporating a minimal set of $500$ transition sequences further consolidates this emergent behavior to $93.8\%$, demonstrating that explicit view conditioning establishes a receptive interface that is readily steerable and sample-efficient to tune.

\subsection{Joint Action-Camera Modeling}
\label{sec:action_camera}

While existing interactive models typically rely on camera trajectories as the primary control signal \citep{gao2026infinite, wu2026infinite, team2026alayaworld} for all view modes, character actuation and camera viewpoint constitute separate degrees of freedom, particularly in \textsc{Tpp}. 
Rather than mapping them to a single control interface \citep{jiang2026abot} or routing them into disparate conditioning branches \citep{sun2025worldplay}, we model character actions and camera trajectories as coexisting, decoupled conditions, contextualizing both directly within the shared transformer sequence.

Given an action sequence and a camera trajectory, \ourmethod encodes both modalities into time-aligned context tokens. For action conditioning, each input $\mathbf a_t$ is decomposed into semantic subslots $\mathcal S=\{\mathrm{move},\mathrm{turn},\Delta\mathrm{trans},\Delta\mathrm{rot}\}$ separating discrete key states from continuous motion increments. The corresponding action tokens are given by:
\begin{equation}
\begin{split}
\mathbf c_{t,p,s}^{\mathrm{action}} = & \operatorname{LayerNorm}\Big( h_s(\mathbf a_t) +r_s(\mathbf a_t) +\mathbf e_t(t) \\ & +\mathbf e_p(p) +\mathbf e_s(s) \Big) +\mathbf e_{\mathrm{type}}(\mathrm{act}),
\end{split}
\end{equation}
where $s\in\mathcal S$, $p\in\mathcal P_\mathrm{act}$ indexes the within-step tokens, $h_s$ denotes a slot-specific encoder (\eg, MLP), and $r_s$ is a zero-initialized residual branch for continuous magnitude cues.

For camera conditioning, \ourmethod converts each normalized relative pose $\widehat{\mathbf E}_t$ into a tokenized Plücker ray field, following \citep{he2025cameractrl, ju2025fulldit}.
For a sampled ray $(u,v)$ with camera center $\widehat{\mathbf o}_t$ and unit direction $\mathbf d_{t,u,v}$, its Plücker coordinate is denoted by $\boldsymbol{\pi}_{t,u,v}=[\widehat{\mathbf o}_t\times\mathbf d_{t,u,v};\mathbf d_{t,u,v}]$. The corresponding camera token is encoded as:
\begin{equation}
\begin{split}
\mathbf c_{t,u,v}^{\mathrm{camera}} = & \operatorname{LayerNorm}\Big(\phi_{\mathrm{ray}}(\boldsymbol{\pi}_{t,u,v}) +\\ &\phi_{\mathrm{pose}}(\operatorname{vec}(\widehat{\mathbf E}_t))  +\phi_{\Delta}(\boldsymbol{\xi}_t) \Big) +\mathbf e_{\mathrm{type}}(\mathrm{cam}),
\end{split}
\end{equation}
where $\boldsymbol{\xi}_t\in\mathbb R^6$ denotes the local rigid-body increment, and the three projections encode ray geometry, global relative pose, and local camera motion, respectively.

Beyond their essential decoupling in \textsc{Tpp}, we further investigate the less-examined \textsc{Fpp} and \textsc{Free} regimes, where camera paths alone are typically assumed to suffice. 
\Cref{fig:43} compares action-only, camera-only, and joint training under our setting on the WBench \textsc{Navigation} set (excluding \textsc{Tpp} scenes). Our key observations are:


\begin{figure}[!h]
\centering
\vspace{-0.8em}
\includegraphics[width=\linewidth]{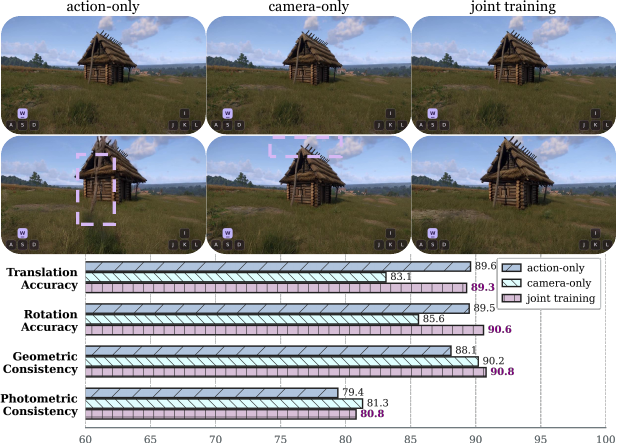}
\vspace{-2em}
\caption{\textbf{Complementary action-camera control.} (\textbf{\textit{Top}}) Action-only training exhibits artifacts; camera-only training yields insufficient displacement. (\textbf{\textit{Bottom}}) Joint training balances motion accuracy and consistency on WBench \textsc{Navigation} (excluding \textsc{Tpp}).}
\label{fig:43}
\vspace{-0.6em}
\end{figure}

\begingroup
\setlength{\tabcolsep}{4.6pt}
\begin{table*}[!t]
\vspace{-1em}
\renewcommand{\arraystretch}{1.34}
  \centering
  \caption{Evaluation on WBench \citep{ying2026wbench} \textsc{Navigation} and \textsc{Fpp$\leftrightarrow$Tpp} sets. We highlight the \textbf{best} and \underline{second best}.}
  \centering
  \vspace{-0.7em}
  \scriptsize
   \begin{tabular}{l ccc|cccccc c}
     \hlineB{2.5}
     \multirow{2}{*}{\textbf{Model}} & \multirow{2}{*}{\textbf{Size}} & \textbf{Supported} & \textbf{Conditioning}  & \multicolumn{6}{c}{\textbf{Navigation}} & \textbf{Perspective} \\
      \cline{5-10}
     &  & \textbf{Controls} & \textbf{Strategy} & \textbf{Quality} & \textbf{Setting} & \textbf{Interaction} & \textbf{Consistency} & \textbf{Physical} & \textbf{Average} &  \textbf{Switching} \\
     \hlineB{1.5}
     EchoWM \citep{zhang2026echowm} & 22B & \actionicon & Positional Attn. &  81.5 & 79.4 & 86.6 & 89.8 & 70.6 & 81.6 & 32.0  \\
     LingBot-World v2 \citep{gao2026infinite} & 14B & \cameraicon & AdaLN & 81.8 & 76.8 & 82.8 & 86.5 & 69.1 & 79.4 & 21.9 \\
     Alaya-EVOKE \citep{yin2026alaya} & 14B & \cameraicon & Warp Condition &  82.8 & 83.8 & 78.6 & 86.9 & 72.1 & 80.8 & 18.8 \\
     HY-World 1.5 \citep{hyworld2025} & 8.3B & \cameraicon & Modulat.+PRoPE & 78.1 & 72.2 & 86.8 & 86.9 & 66.3 & 78.1 & 12.5  \\
     \hline
     Wan 2.2 \citep{wan2025wan} & 5B & Text &  & 72.2 & 69.3 & 65.4 & 86.4 & 52.3 & 69.1 & 6.3 \\
     YUME 1.5 \citep{mao2026yume1} & 5B & Text & Cross-Attn. & 77.6 & 72.4 & 71.4 & 80.1 & \underline{65.2} & 73.3 & 12.5  \\
     ABot-World \citep{jiang2026abot} & 5B & \actionicon~\cameraicon & Additive Inject. & 76.8 & 71.4 & 84.0 & 79.5 & 61.7 & 74.7 & 9.4 \\
     DreamX-World \citep{team2026dreamx} & 5B & \cameraicon & Positional Attn. & 77.5 & \underline{80.8} & 78.6 & 74.9 & 63.3 & 75.0 & \underline{18.8} \\
     Zing-0.5 \citep{chen2026zing} & 5B & \actionicon & Additive Inject. & \underline{80.6} & 77.8 & \underline{84.2} & \textbf{88.5} & \textbf{73.8} & \underline{81.0} & 12.5  \\
     \rowcolor{blue!10}
     \textbf{\ourmethod (Ours)} & 5B & \actionicon~\cameraicon & In-Context & \textbf{80.9} & \textbf{81.0} & \textbf{86.0} & \underline{87.2} & \underline{72.3} & \textbf{81.5} & \textbf{93.8} \\
     \hlineB{2.5}
   \end{tabular}
  \label{tab:main}
  \vspace{-1em}
\end{table*} 
\endgroup

\noindent\textbf{Obs.\ding{184} Action and camera signals play complementary roles.}\quad As shown in \Cref{fig:43}, action-only conditioning yields higher translation and rotation accuracy ($+6.5$ and $+3.9$ points over camera-only), but can introduce visual artifacts. 
Conversely, camera-only conditioning improves geometric ($+2.1$) and photometric ($+1.9$) consistency, though often at the cost of insufficient travel displacement. 
Joint training naturally harmonizes these properties, achieving the highest rotation accuracy ($90.6$) and geometric consistency ($90.8$) while maintaining strong translation fidelity ($89.3$). 
In effect, action commands govern local responsiveness, whereas camera trajectories anchor global stability.

\subsection{Interaction-Transport Geometric Encoding}
\label{sec:rope}

While in-context tokens expose control inputs to the shared attention space, standard 3D-RoPE encodes only relative spatiotemporal coordinates, leaving the intervening control progression between distant frames implicit. 
To this end, \ourmethod further introduces Interaction-Transport RoPE (IT-RoPE), which complements spatiotemporal positional encoding by parameterizing temporal attention with control-induced transport operators accumulated along the execution sequence.

\vspace{-1em}
\paragraph{Ordered Interval Transport.}
While input-dependent transforms \citep{yang2025path} and geometric encoding \citep{li2025cameras} demonstrate the promise of data-driven attention biases, our interactive setting differs in that cross-temporal transitions are governed by temporally ordered user inputs. 
We thus maintain \textit{dual, order-preserving transport paths} for character action and camera motion $b \in \{a, c\}$.
Specifically, each control input induces a step transformation $\mathbf{U}_t^b$, which accumulates along continuous valid segments to define the relative transport across interval $(i, j]$:
\begin{equation}
\mathbf{T}_{i \rightarrow j}^b = \prod_{k = i + 1}^{j} \mathbf{U}_k^b, \quad i < j,
\label{eq:interval_transport}
\end{equation}
where $\mathbf{T}_{j \rightarrow i}^b = (\mathbf{T}_{i \rightarrow j}^b)^\top$, and $\mathbf{T}_{i \rightarrow j}^b \mathbf{T}_{j \rightarrow k}^b = \mathbf{T}_{i \rightarrow k}^b$. 
This non-commutative composition natively preserves control order, allowing the model to distinguish trajectories that share identical endpoints but differ in execution sequences.

\vspace{-1em}
\paragraph{View-Gated Centered Attention.}
To incorporate these accumulated transitions into cross-frame reasoning, we augment temporal attention with a centered, transport-corrected score:
\begin{equation}
s_{ij} = s_{ij}^{\text{base}} + \sum_{b \in \{a, c\}} \frac{\lambda^b}{\sqrt{r}} m_{ij}^b g_i^b g_j^b \cdot (\mathbf{q}_i^b)^\top \left( \mathbf{T}_{t_i \rightarrow t_j}^b - \mathbf{I} \right) \mathbf{k}_j^b,
\label{eq:it_rope_attn}
\end{equation}
where $s_{ij}^{\text{base}} = \frac{\bar{\mathbf{q}}_i^\top \bar{\mathbf{k}}_j}{\sqrt{d}}$ denotes standard spatiotemporal attention with 3D-RoPE, $\mathbf{q}_i^b, \mathbf{k}_j^b \in \mathbb{R}^r$ are control-specific sub-channels, and $\lambda^b$ is a learnable scale initialized to zero. 
The gate $g_i^b \in (0, 1)$ dynamically modulates control influence across viewing regimes and token types, while $m_{ij}^b \in \{0, 1\}$ restricts the correction to contiguous valid segments. 
Crucially, subtracting $\mathbf{I}$ centers the correction, ensuring it vanishes for static intervals or $t_i = t_j$, while zero-initializing $\lambda^b$ preserves native backbone behavior at the start of training. 
Further details of transport accumulation and attention computation are provided in \Cref{app_sec:it_rope_details}.

\subsection{Training of World-in-Context}
\label{sec:training}

\paragraph{Training Data.}
We train \ourmethod on a compact set of $100$K video clips covering \textsc{Free}, \textsc{Fpp}, and \textsc{Tpp} viewing regimes, accompanied by temporally aligned view, action, and camera annotations when available. 
While joint training across these single-regime clips naturally facilitates view-mode transitions (as analyzed in \Cref{sec:view_mode}), we additionally incorporate a small fraction of cross-view switching clips ($500$ sequences) with dynamic view schedules to reinforce transition stability. 

\vspace{-1em}
\paragraph{Training Protocol.}
Following the conditional flow-matching objective, \ourmethod undergoes bidirectional finetuning where video and control tokens jointly interact in the shared attention manifold. 
This bidirectional setup serves as our default configuration to benchmark the core in-context conditioning paradigm. 
In \Cref{sec:applicability}, we further investigate \ourmethod under broader operational settings, including \textit{causal rollout with in-context history} and \textit{expanded game-control signals}.
\section{Experiments}
\label{sec:experiments}

\begin{figure*}[!t]
\centering
\includegraphics[width=\linewidth]{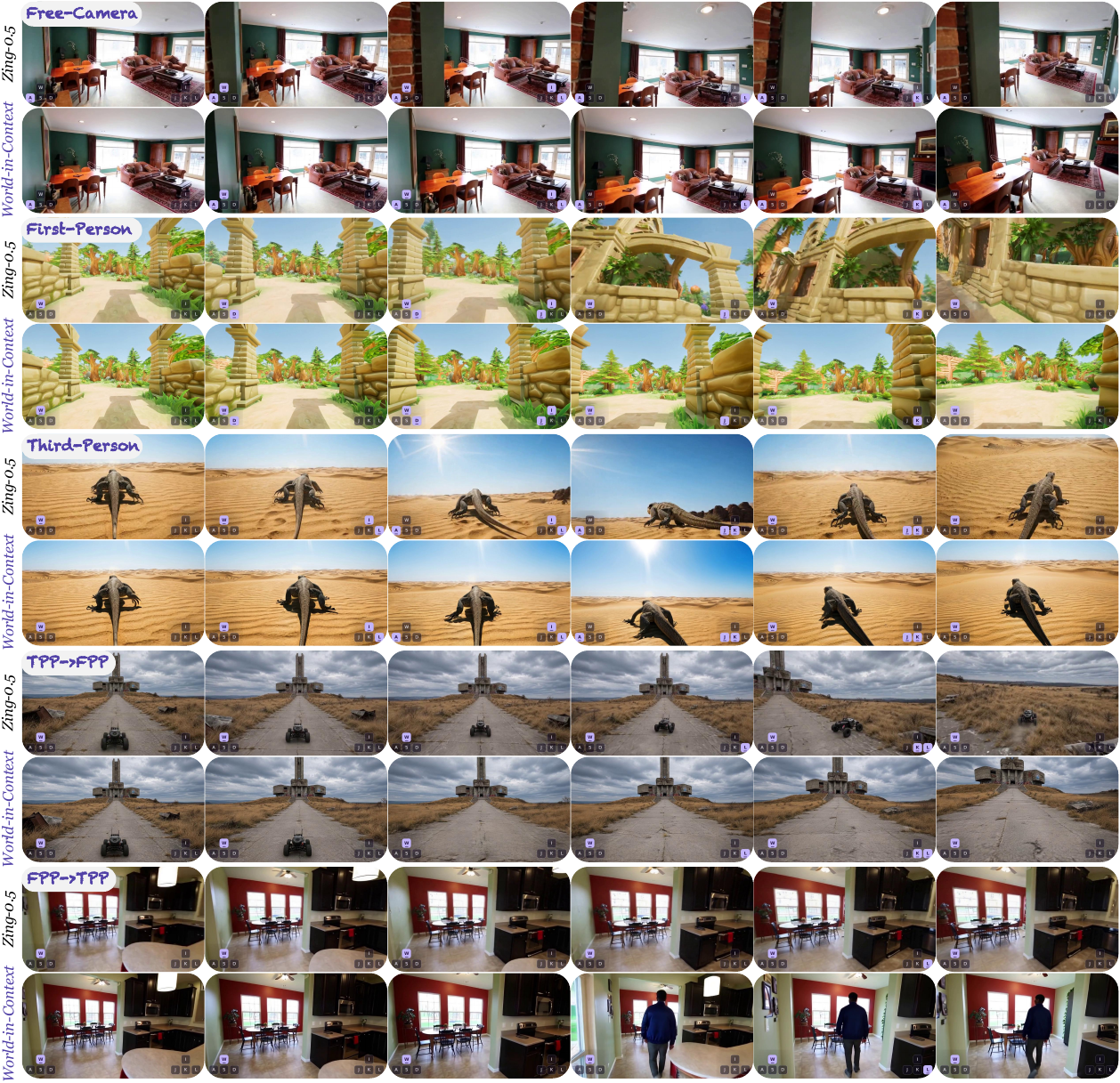}
\vspace{-2em}
\caption{Qualitative comparisons of \ourmethod.}
\label{fig:qualitative}
\vspace{-0.6em}
\end{figure*}

\subsection{Experimental Settings}

\paragraph{Baselines.} \ourmethod is built upon the widely adopted Wan2.2-TI2V-5B \citep{wan2025wan} backbone. 
For direct comparisons, our primary baselines comprise models sharing this foundation: YUME~1.5 \citep{mao2026yume1}, ABot-World \citep{jiang2026abot}, DreamX-World \citep{team2026dreamx}, and Zing-0.5 \citep{chen2026zing}. 
We also include larger models as reference: EchoWM \citep{zhang2026echowm}, LingBot-World~v2 \citep{gao2026infinite}, Alaya-EVOKE \citep{team2026alayaworld}, and HY-World~1.5 \citep{hyworld2025}. 
Since training data scales are rarely disclosed across works, we also implement public architectures and conduct controlled comparisons under our identical data for analyses.

\vspace{-1em}
\paragraph{Evaluations.} We evaluate \ourmethod on the widely adopted WBench \citep{ying2026wbench}. 
Alongside the standard \textsc{Navigation} split for trajectory tracking, we further adopt its \textsc{Perspective Switching - Fpp$\leftrightarrow$Tpp} suite to systematically assess dynamic view transitions. Our framework natively supports both action-only and camera-only inference; we default to action-only control for \textsc{Free} and \textsc{Fpp}, while utilizing joint action and camera signals for character and viewpoint in \textsc{Tpp}. 
More details are in \Cref{app_sec:experiemntal_details}.

\begin{figure*}[!t]
\centering
\vspace{1em}
\includegraphics[width=\linewidth]{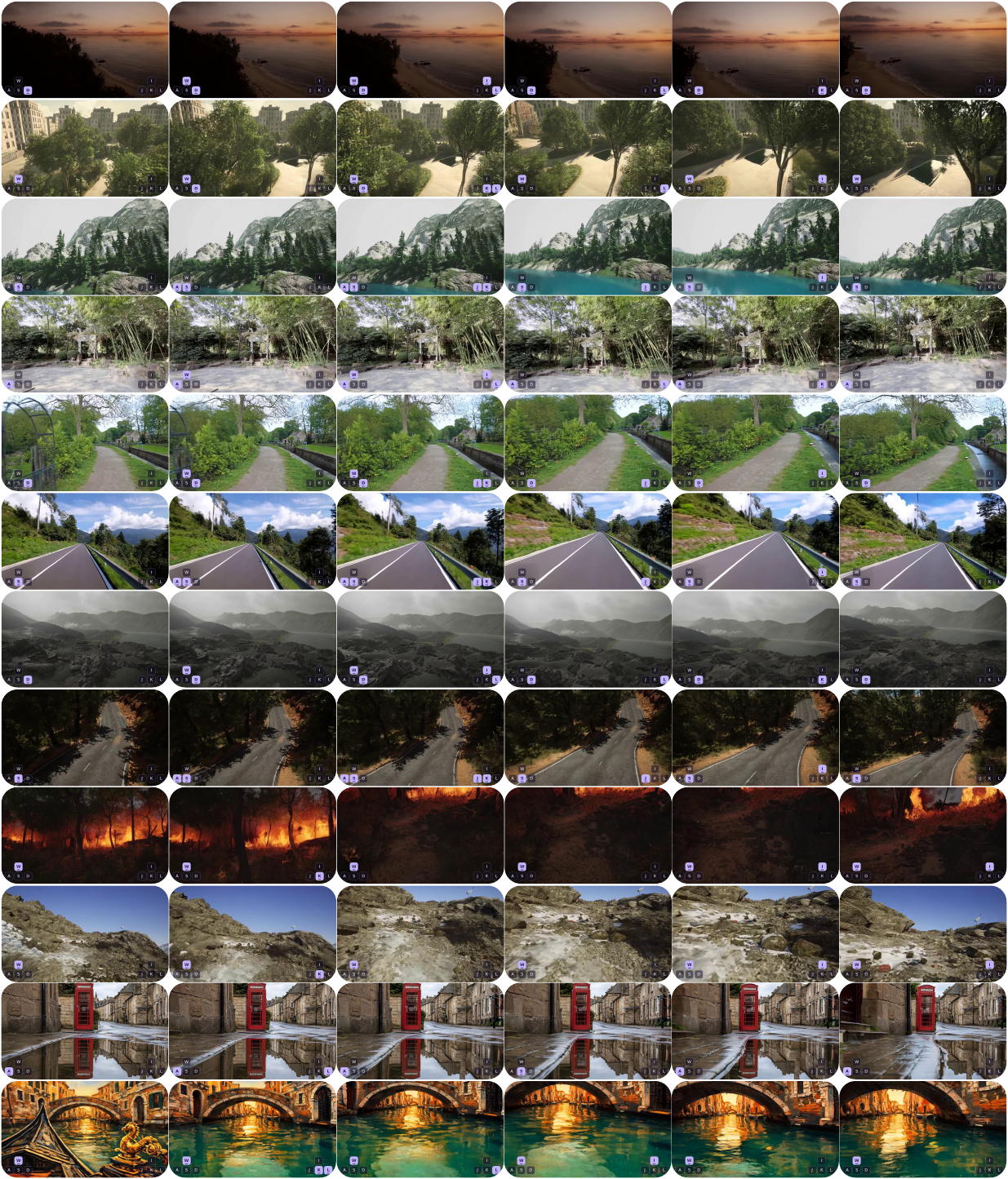}
\vspace{-1.6em}
\caption{Qualitative results of \ourmethod on \textsc{Free} mode.}
\label{app_fig:qualitative_free}
\end{figure*}

\begin{figure*}[!t]
\centering
\vspace{1em}
\includegraphics[width=\linewidth]{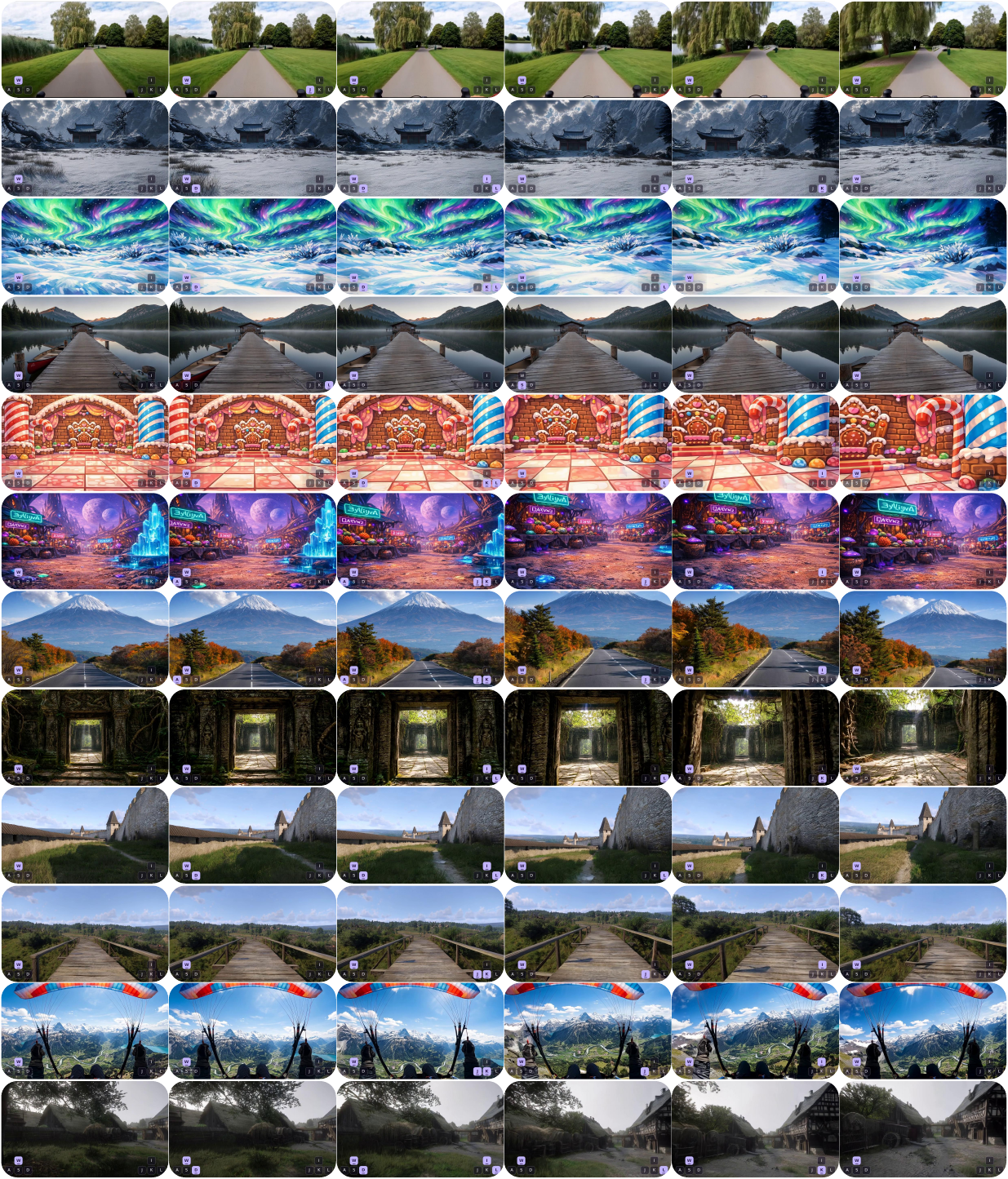}
\vspace{-1.6em}
\caption{Qualitative results of \ourmethod on \textsc{Fpp} mode.}
\label{app_fig:qualitative_fpp}
\end{figure*}

\begin{figure*}[!t]
\centering
\vspace{1em}
\includegraphics[width=\linewidth]{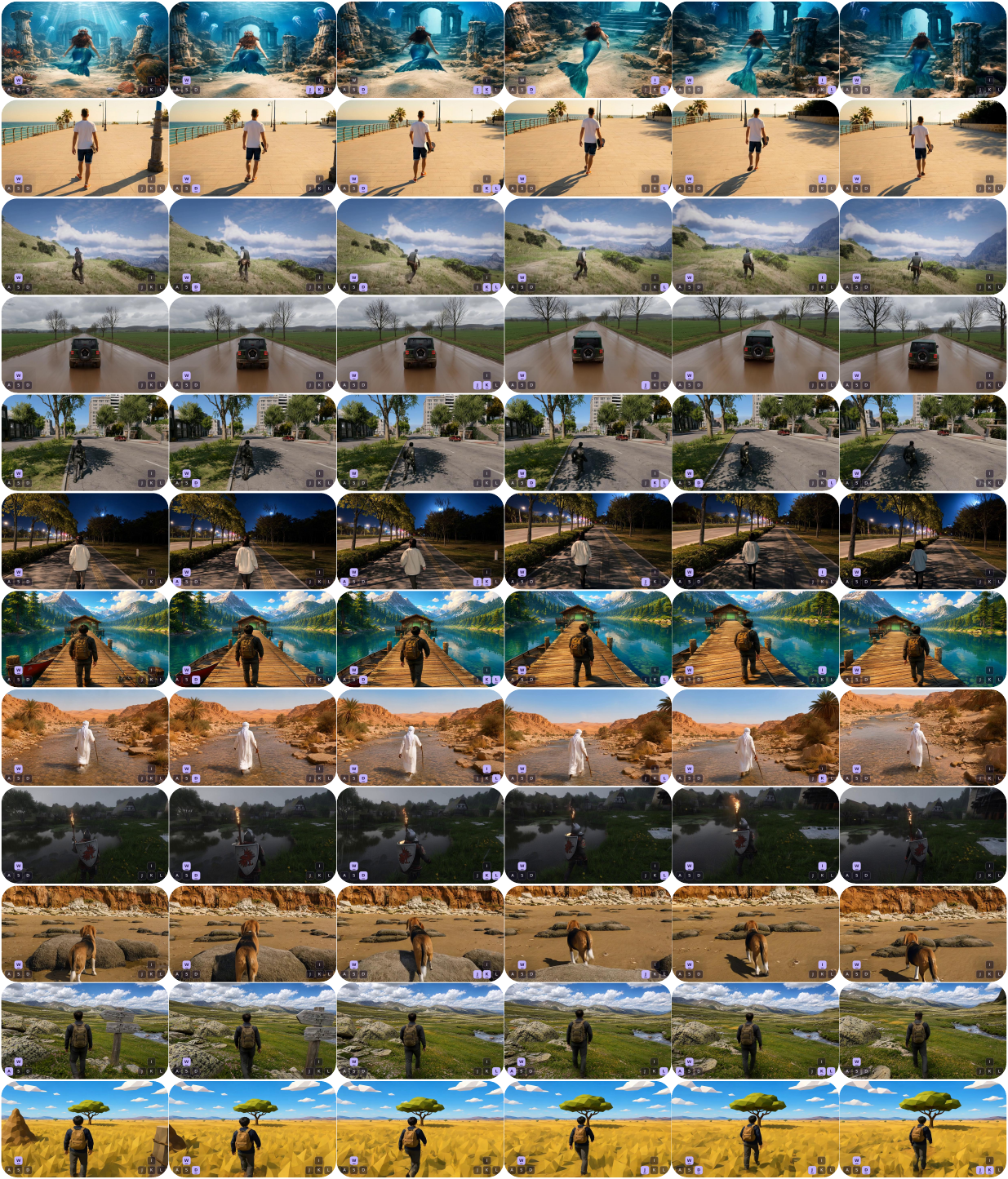}
\vspace{-1.6em}
\caption{Qualitative results of \ourmethod on \textsc{Tpp} mode.}
\label{app_fig:qualitative_tpp}
\end{figure*}

\begin{figure*}[!t]
\centering
\vspace{1em}
\includegraphics[width=\linewidth]{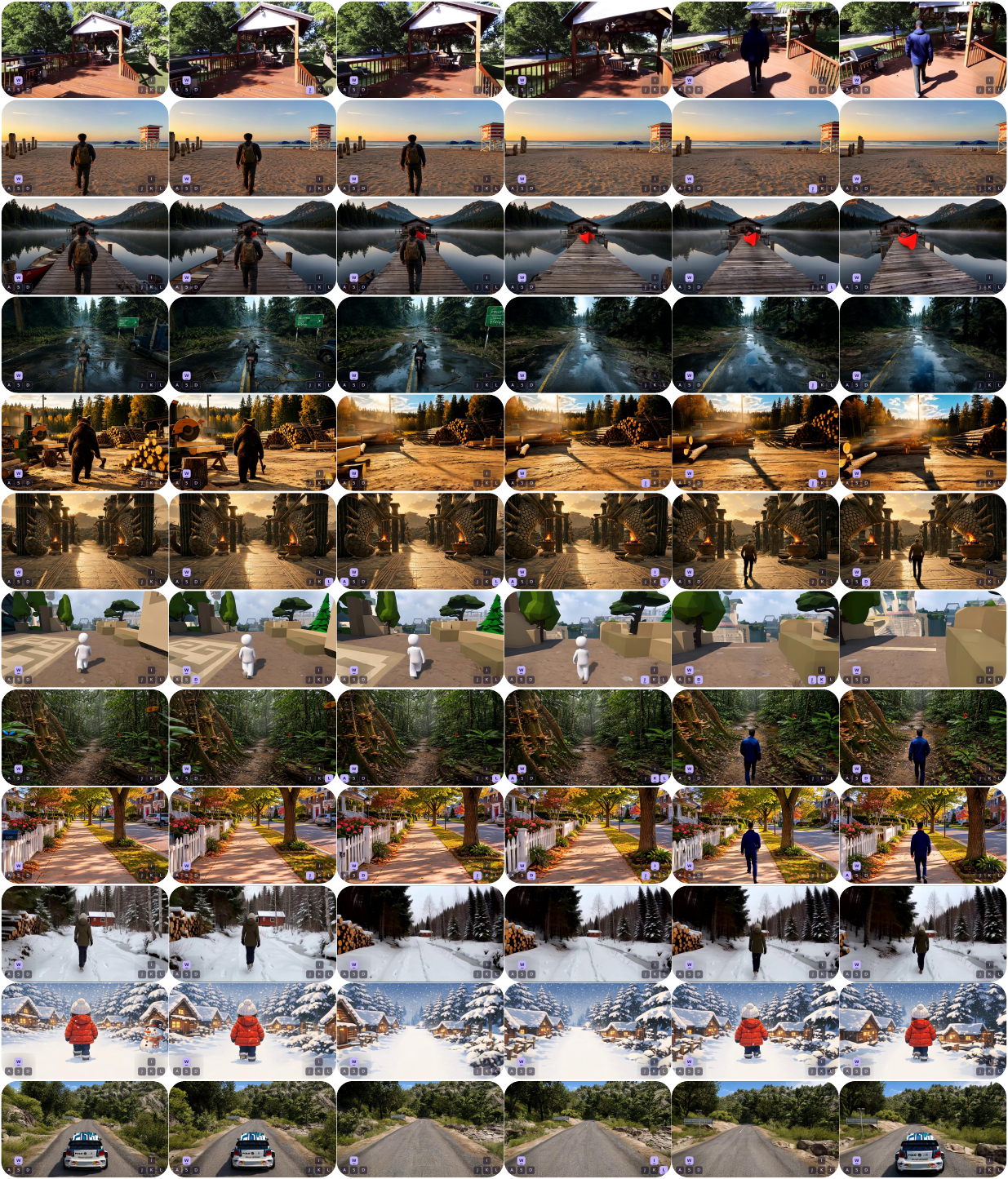}
\vspace{-1.6em}
\caption{Qualitative results of \ourmethod on \textsc{Fpp} $\leftrightarrow$ \textsc{Tpp}.}
\label{app_fig:qualitative_switch}
\end{figure*}

\subsection{Main Results}

\paragraph{Quantitative Results.}
\Cref{tab:main} summarizes performance on WBench. 
Among 5B models, \ourmethod achieves the highest \textsc{Navigation} average ($81.5$), performing on par with Zing-0.5 ($81.0$) while improving the \textsc{Interaction} score from $84.2$ to $86.0$ via unified in-context conditioning. 
Its navigation capability also rivals larger models like 22B EchoWM ($81.6$). 
Crucially, \ourmethod reaches a $93.8\%$ success rate on \textsc{Fpp$\leftrightarrow$Tpp}, significantly outperforming the best 5B baseline ($18.8\%$) and confirming its capacity for dynamic viewpoint transitions within a single unified model.

\vspace{-1em}
\paragraph{Qualitative Results.}
\Cref{fig:qualitative} further compares our \ourmethod with Zing-0.5. 
While both generate plausible environments, Zing-0.5 tends to incur exaggerated motion scale, whereas ours maintains more measured, balanced trajectories aligned with its higher \textsc{Interaction} score. 
Furthermore, \ourmethod smoothly executes \textsc{Fpp}$\leftrightarrow$\textsc{Tpp} switches where the baseline fails to transition. 
Extended rollouts across diverse scene layouts and artistic styles under \textsc{Free}, \textsc{Fpp}, and \textsc{Tpp} modes are detailed in \Cref{app_fig:qualitative_free,app_fig:qualitative_fpp,app_fig:qualitative_tpp}. Notably, \Cref{app_fig:qualitative_switch} also demonstrates coherent multi-stage switching (\eg, \textsc{Tpp}$\rightarrow$\textsc{Fpp}$\rightarrow$\textsc{Tpp}) while preserving consistent scene structure throughout.

\subsection{Analysis of In-Context Conditioning}
\label{sec:5.3}

To further examine how conditioning strategies scale with input complexity, we compare ICC with additive injection \citep{jiang2026abot} and AdaLN modulation \citep{team2026advancing} under progressively richer inputs with our same data: view (V), action (A), and camera (C). We give the following observations:

\begin{figure}[!h]
\centering
\vspace{-0.6em}
\includegraphics[width=\linewidth]{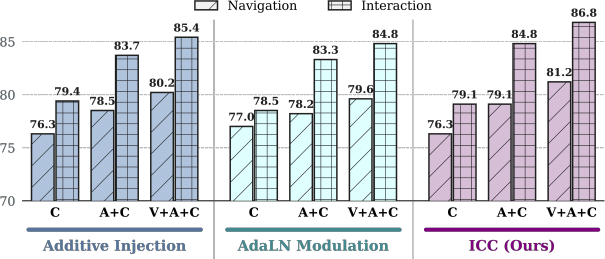}
\vspace{-2.2em}
\caption{\textbf{Conditioning strategy analysis.} ICC increasingly benefits the coordination of diverse controls as conditioning becomes richer. C/A/V denote camera/action/view.}
\label{fig:53}
\vspace{-0.6em}
\end{figure}

\noindent\textbf{Obs.\ding{185} ICC's advantage scales with conditioning richness.}\quad
As shown in \Cref{fig:53}, with camera control alone, ICC performs on par with alternative strategies ($76.3$ \textit{vs.} $76.3$-$77.0$). Its gains emerge under joint action-camera inputs ($+0.6$ in \textsc{Navigation}, $+1.1$ in \textsc{Interaction}) and expand further when adding explicit view tokens ($+1.0$ and $+1.4$, reaching $81.2$ and $86.8$). These widening margins indicate that a shared attention space coordinates diverse controls more effectively than separate injection pathways as interaction becomes compositional.

\subsection{Analysis of IT-RoPE}
\label{sec:5.4}

\begin{figure}[!h]
\centering
\vspace{-0.6em}
\includegraphics[width=\linewidth]{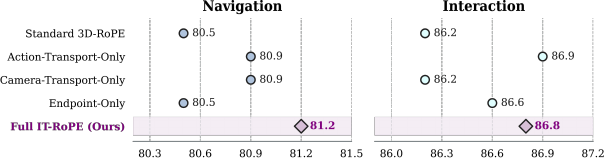}
\vspace{-2em}
\caption{\textbf{IT-RoPE ablation.} Accumulating intervening control transitions improves navigation and control fidelity beyond endpoint-only conditioning.}
\label{fig:54}
\vspace{-0.6em}
\end{figure}

To evaluate how trajectory-aware attention affects temporal consistency, we ablate IT-RoPE here against standard 3D-RoPE, single-branch variants, and an endpoint-only alternative under the fixed V+A+C setup, as shown in \Cref{fig:54}.

\noindent\textbf{Obs.\ding{186} Tracking intervening control evolution improves interactive attention.}\quad 
Compared to 3D-RoPE, full IT-RoPE enhances both \textsc{Navigation} ($80.5\rightarrow 81.2$) and \textsc{Interaction} ($86.2\rightarrow 86.8$). 
While action transport sharpens command responsiveness and camera transport stabilizes spatial tracking, their combination delivers the most balanced performance. 
Moreover, full IT-RoPE clearly outperforms the endpoint-only variant ($80.5/86.6$), confirming that accumulating the intervening control sequence provides critical dynamics missed by endpoint offsets alone.

\subsection{Analysis of Data Efficiency}

Our default interaction-training set comprises a modest $100$K navigation clips and $500$ switching sequences. Since training-data scales are not consistently disclosed by existing methods, we assess data efficiency through controlled comparisons in \Cref{fig:55}: $50$K+$250$, $100$K+$500$, and $125$K+$750$.
We summarize our observations as follows:

\begin{figure}[!h]
\centering
\vspace{-0.6em}
\includegraphics[width=\linewidth]{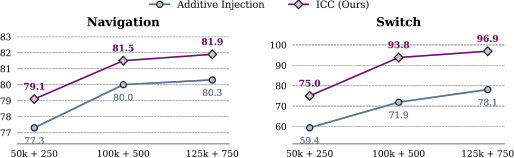}
\vspace{-2em}
\caption{\textbf{Data scaling analysis.} ICC consistently outperforms additive injection across training budgets.}
\label{fig:55}
\vspace{-0.6em}
\end{figure}

\noindent\textbf{Obs.\ding{187} ICC demonstrates data efficiency across scaling regimes.}\quad ICC consistently outperforms additive injection across all data regimes, especially in perspective switching. 
With only $50$K+$250$ clips, ICC achieves a $75.0\%$ switching rate, already surpassing additive injection trained on twice the data ($71.9\%$ at $100$K+$500$). 
At $100$K+$500$, ICC exceeds the largest-budget additive baseline ($125$K+$750$) in both \textsc{Navigation} ($81.5$ \textit{vs.} $80.3$) and \textsc{Switch} ($93.8\%$ \textit{vs.} $78.1\%$), showing that shared in-context tokens make substantially more efficient use of interactive supervision.

\begin{figure*}[!t]
\centering
\vspace{-0.4em}
\includegraphics[width=\linewidth]{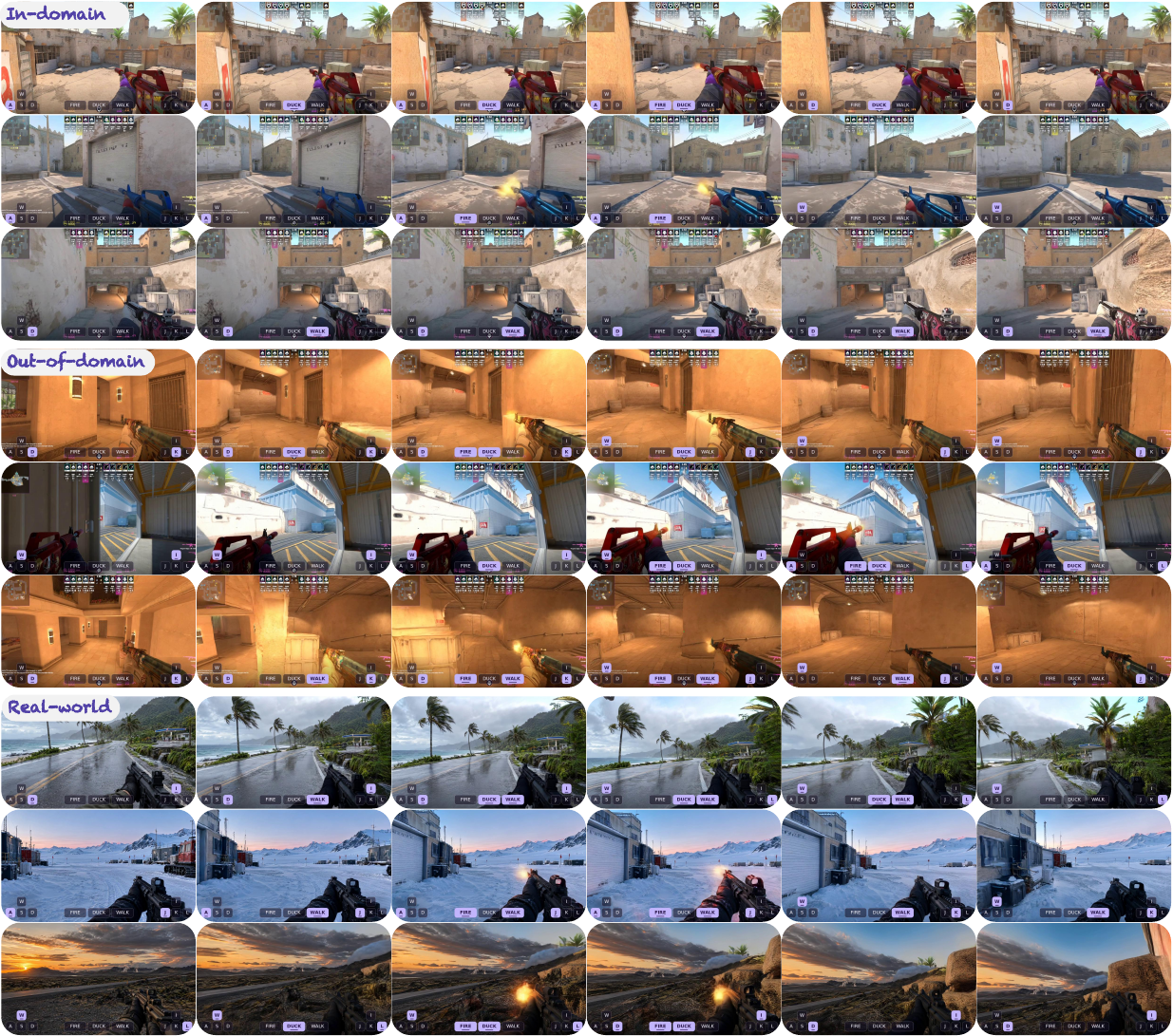}
\vspace{-1.6em}
\caption{Qualitative results of \ourgame.}
\label{app_fig:qualitative_game_controls}
\end{figure*}

\section{Applicability of World-in-Context}
\label{sec:applicability}

\subsection{Causal Modeling with In-Context History}

To examine the architectural extensibility of in-context conditioning beyond bidirectional generation, \ourmethod is naturally adapted to streaming causal rollouts. Following common causal training paradigms in interactive generation \citep{zhu2026causal}, the model is trained autoregressively to predict future frame latents conditioned on historical context. Rather than relying on external KV cache retrieval mechanisms, \ourcausal directly incorporates historical latents into the shared self-attention sequence. Specifically, the condition sequence $\mathbf{c}$ in Eq.~\eqref{eq:unified_condition} is augmented with historical latent tokens:
\begin{equation}
    \mathbf{c}_\mathrm{causal} = \big[ \mathbf{c}^{\mathrm{view}} \,;\, \mathbf{c}^{\mathrm{action}} \,;\, \mathbf{c}^{\mathrm{camera}} \,;\, \mathbf{c}^{\mathrm{sink}} \,;\, \mathbf{c}^{\mathrm{recent}} \big],
\end{equation}
where $\mathbf{c}^{\mathrm{sink}}$ preserves initial context tokens as global semantic anchors, and $\mathbf{c}^{\mathrm{recent}}$ maintains a sliding window of recent visual tokens to enforce continuity.

\begin{table}[!h]
\centering
\vspace{-0.6em}
  \caption{Evaluation of \ourcausal on WBench.}
    \vspace{-1em}
    \renewcommand{\arraystretch}{1.3}
    \footnotesize
    \resizebox{0.48\textwidth}{!}{%
   \begin{tabular}{lccccccc}
     \hlineB{2.5}
    \rowcolor{CadetBlue!20} 
    \textbf{Method} & \textbf{Qual.} & \textbf{Set.} & \textbf{Inter.} & \textbf{Consi.} & \textbf{Phys.} & \textbf{Avg.} & \textbf{Switch.} \\
    \hline
     Wan 2.2 \citep{wan2025wan} & 72.2 & 69.3 & 65.4 & 86.4 & 52.3 & 69.1 & 6.3 \\
     \rowcolor{gray!10}
     YUME 1.5 \citep{mao2026yume1} &  77.6 & 72.4 & 71.4 & 80.1 & 65.2 & 73.3 & 12.5  \\
     ABot-World \citep{jiang2026abot} & 76.8 & 71.4 & 84.0 & 79.5 & 61.7 & 74.7 & 9.4 \\
     \rowcolor{gray!10}
     DreamX-World \citep{team2026dreamx} & 77.5 & 80.8 & 78.6 & 74.9 & 63.3 & 75.0 & 18.8 \\
     Zing-0.5 \citep{chen2026zing} & 80.6 & 77.8 & 84.2 & 88.5 & 73.8 & 81.0 & 12.5  \\
     \hline
     \rowcolor{gray!10}
     \textbf{\ourmethod} & 80.9 & 81.0 & 86.0 & 87.2 & 72.3 & 81.5 & 93.8 \\
     \textbf{\ourcausal} & 80.5 & 80.4 & 85.7 & 87.6 & 72.0 & 81.2 & 90.6 \\
     \hlineB{2.5}
   \end{tabular}%
}
  \label{tab:causal_wbench}
  \vspace{0.8em}
\end{table} 

\begin{figure*}[!t]
\centering
\vspace{0.4em}
\includegraphics[width=\linewidth]{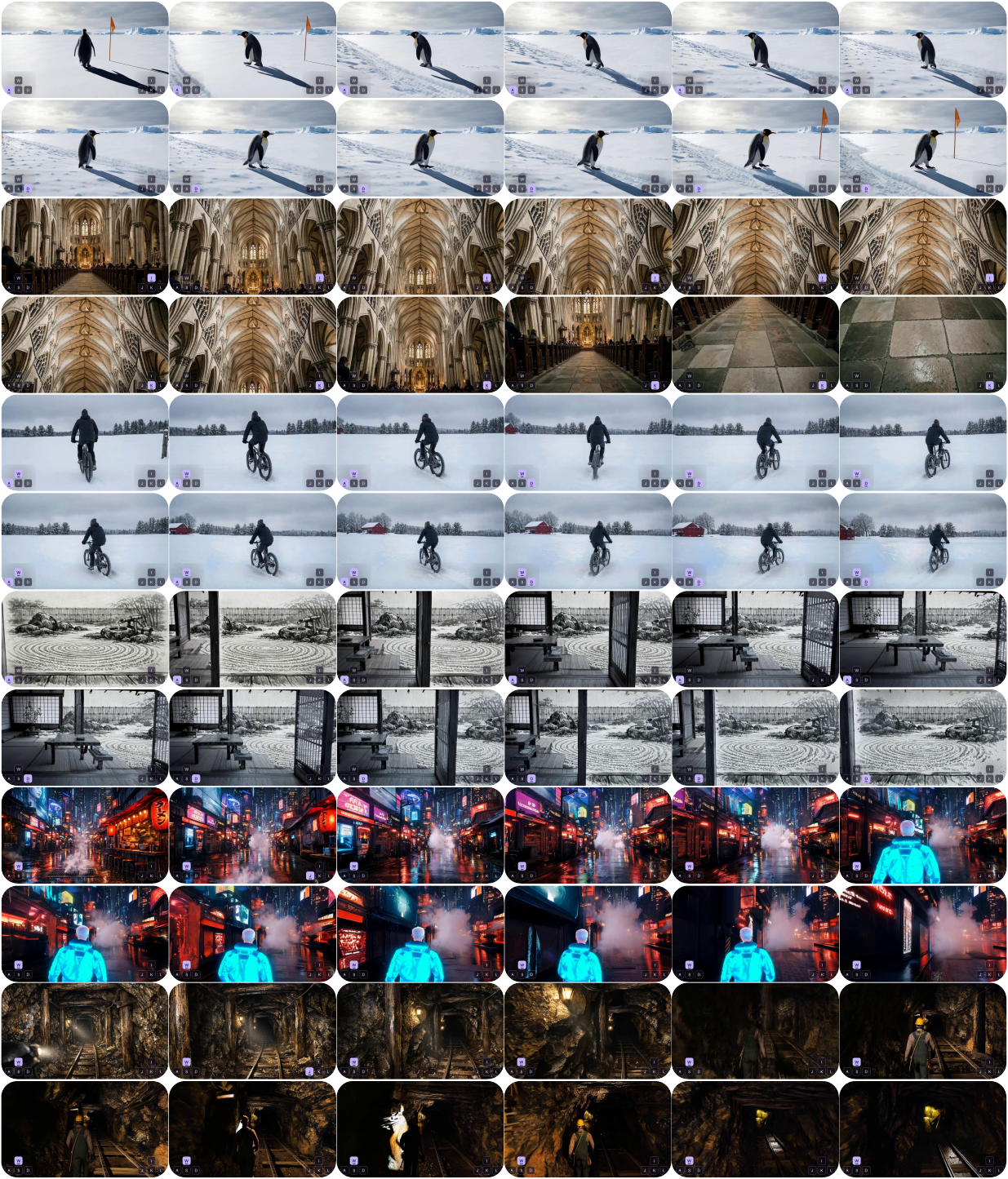}
\vspace{-1.6em}
\caption{Qualitative results of \ourcausal.}
\label{app_fig:qualitative_causal}
\vspace{1em}
\end{figure*}

As shown in \Cref{tab:causal_wbench}, \ourcausal exhibits competitive interactive performance, attaining an average \textsc{Navigation} score of $81.2$ and a $90.6\%$ \textsc{Fpp$\leftrightarrow$Tpp} success rate on WBench. Compared to the bidirectional counterpart ($81.5$ and $93.8\%$), causal modeling incurs negligible degradation while retaining a better performance margin over existing 5B baselines. Furthermore, qualitative rollouts in \Cref{app_fig:qualitative_causal} demonstrate that \ourcausal preserves long-horizon spatial coherence across diverse viewing regimes. Notably, in loop-closure and viewpoint reappearance scenarios (\eg, returning to previously explored landscapes or room corners), \ourcausal consistently reproduces previously observed visual landmarks and layouts.

\subsection{Scaling to Rich Game Controls}

To further demonstrate the extensibility and token-level composability beyond canonical navigation commands, we extend \ourmethod to accommodate richer, high-frequency gameplay actions. Experiments are conducted on Counter-Strike 2 gameplay sequences from EWM \citep{ma2026esports}. To assess sample efficiency and generalizability, \ourgame is trained on a compact subset of a single map (\ie, \textsc{Dust II}), and tested on unseen layouts as well as real-world scenarios. Specifically, the condition sequence is augmented with discrete game-control tokens:
\begin{equation}
    \big[ \mathbf{c}^{\mathrm{fire}} \,;\, \mathbf{c}^{\mathrm{duck}} \,;\, \mathbf{c}^{\mathrm{walk}} \big],
\end{equation}
corresponding to firing (weapon dynamics and muzzle flashes), crouching (viewport descent), and slow walking (motion attenuation), respectively.

As illustrated in \Cref{app_fig:qualitative_game_controls}, \ourgame demonstrates sensitive, decoupled control execution across three complementary regimes:
(\textbf{\textit{i}}) \textit{In-domain}: new control trajectories within the training map (\textsc{Dust II}) exhibit instant actuation responses;
(\textbf{\textit{ii}}) \textit{Out-of-distribution}: when applied to unseen game maps (\eg, \textsc{Mirage}, \textsc{Nuke}), the model preserves scene consistency while faithfully following concurrent actions;
(\textbf{\textit{iii}}) \textit{Real-world transfer}: using synthetic starting images, the learned actuation dynamics generalize naturally to realistic visual environments.

\section{Conclusion and Positioning}

\paragraph{Conclusion.} In this work, we present \ourmethod, a video world modeling framework based on In-Context Conditioning (ICC) that unifies view mode, character action, and camera trajectory as tokens within a shared attention space. 
We systematically investigate core interactive dimensions, revealing that explicit view tokens disentangle conflicting motion dynamics while enabling emergent in-video switching, and that action and camera signals complement each other in local responsiveness and global stability. 
Augmented with IT-RoPE to encode accumulated control progression in temporal attention, \ourmethod achieves competitive navigation on WBench ($81.5$) matching the 22B EchoWM, while establishing strong perspective-switching capability ($93.8\%$). 
Controlled analyses further demonstrate compounding advantages under richer conditioning, favorable sample efficiency, and natural applicability to causal rollouts, highlighting shared context modeling as a scalable interface for interactive world simulation.

\vspace{-1em}
\paragraph{Positioning of This Report.}
This work focuses primarily on the algorithmic formulation of interactive conditioning, rather than training an all-encompassing system artifact, with \ourmethod serving as a concrete instantiation of this design. The overall capacity of an interactive world model naturally depends on data quality, model size, and training scale. We use a modest interaction-training set of $100$K navigation clips and $500$ switching sequences. Since comparable training-data statistics are not consistently disclosed by existing methods, we omit cross-method data-scale comparisons from the main table. Instead, we prioritize controlled comparisons that isolate the architectural advantages of shared context modeling and control-aware temporal attention, including their effectiveness across training-data budgets.
Specifically, this work centers around three foundational control modalities, \textit{i.e.}, view mode, character action, and camera motion, as they directly address the core dimensions governing interactive navigation: how the environment is viewed across perspectives and how user inputs steer character actuation alongside viewport evolution. These algorithmic findings complement concurrent advances in scaling, presenting an extensible interface for coordinating diverse interactive controls.

\vspace{0.8em}
\section*{Contributions}

\paragraph{Contributors.}
Harold Haodong Chen$^*$, Rui Peng$^\dagger$,

Rongjin Guo,~~Lingteng Qiu,~~Kejie Qiu,~~Chao Xu,

Zilong Dong

\vspace{1em}
$^\dagger$Corresponding Author

$^*$Work done during Harold H. Chen's internship at ATH@Alibaba.

\vspace{1em}
For questions or feedback, feel free to contact \href{mailto:haroldchen328@gmail.com}{haroldchen328@gmail.com} or open a \href{https://github.com/aigc3d/World-in-Context}{GitHub} issue.

\newpage 
{
    \small
    \bibliographystyle{ieeenat_fullname}
    \bibliography{main}
}

\clearpage
\appendix
\setcounter{page}{1}

\section{Training Details of \ourmethod}
\label{app_sec:training_details}

\subsection{Training Data}

\paragraph{Data Sources.} The training mixture comprises a default collection of $100$K single-regime navigation clips alongside a subset of $500$ \textsc{Fpp$\leftrightarrow$Tpp} switching sequences. \textit{All qualitative evaluation scenes presented in this work are strictly held out from the training set.}
\vspace{0.2em}
\begin{itemize}[leftmargin=1.4em]
    \item[\ding{111}] \textit{Navigation Data ($100$K)}: Clips are collected across widely-adapted open-source video datasets, including DL3DV \citep{ling2024dl3dv}, RealEstate \citep{zhou2018stereo}, Sekai \citep{li2026sekai}, SpatialViD \citep{wang2026spatialvid}, EgoExo \citep{song2026worldwander}, and ABot-World-Explorer \citep{jiang2026abot}.
    \item[\ding{111}] \textit{Switching Data ($500$)}: Viewpoint-transition sequences are collected from EgoExo \citep{song2026worldwander} and internal data.
\end{itemize}

\vspace{-1em}
\paragraph{Quality Pipeline.} The raw video corpus is filtered through a standard three-stage automated filtering pipeline: 
\begin{itemize}[leftmargin=1.4em]
    \item[\ding{111}] \textit{Scene/Shot Continuity}: Following LingBot-World \citep{team2026advancing}, TransNetV2 \citep{soucek2024transnet} is deployed to detect abrupt scene transitions and shot boundaries; segmented multi-shot videos are removed to retain strictly continuous navigation segments.
    \item[\ding{111}] \textit{Motion Filtering}: Following Open-Sora \citep{zheng2024open} and SANA-Video \citep{chen2026sana}, UniMatch \citep{yang2025unimatch} optical flow and VMAF metrics are evaluated to quantify dynamic motion magnitude. Clips displaying static stagnation or excessive motion blur/discontinuities are discarded.
    \item[\ding{111}] \textit{Aesthetic Assessment}: Dover \citep{wu2023exploring} is utilized to score perceptual aesthetics, eliminating clips compromised by severe compression artifacts, overexposure, or low visual fidelity.
\end{itemize}

\vspace{-1em}
\paragraph{Control Annotation.} The control signals mainly used in \ourmethod, \textit{i.e.}, view, camera, and action, are annotated as follows:
\vspace{0.2em}
\begin{itemize}[leftmargin=1.4em]
    \item[\ding{111}] \textit{View Mode}: A vision-language foundation model (\textit{i.e.}, Qwen3-VL-32B \citep{bai2025qwen3}) classifies clips into \textsc{Free}, \textsc{Fpp}, and \textsc{Tpp} regimes. \textsc{Free} is characterized by unconstrained, smooth aerial or camera roaming (\eg, drone cinematography); \textsc{Fpp} captures embodied locomotion or in-cockpit perspectives; \textsc{Tpp} encompasses external character-following camera arrangements. Ambiguous instances at boundaries are cross-verified and resolved using reconstructed camera trajectory statistics.
    \item[\ding{111}] \textit{Camera Trajectories}: Following common practice, continuous 6-DoF camera trajectories and Plücker ray parameters are reconstructed using ViPE \citep{huang2025vipe}.
    \item[\ding{111}] \textit{Action Inversion}: While authentic character actuation natively originates from game-engine telemetry, obtaining massive synchronized game recordings remains challenging. Therefore, discrete character actions are mainly inversely resolved from ViPE-estimated continuous camera trajectories. Crucially, for \textsc{Fpp} sequences, body-induced periodic oscillations and head-bobbing motions are decoupled and filtered out before discrete action discretization.
\end{itemize}

\vspace{-1em}
\paragraph{World-State Captioning.} Unlike existing world-modeling frameworks (\eg, LingBot-World \citep{team2026advancing}, Matrix-Game 3.0 \citep{wang2026matrix}) that synthesize verbose temporal narratives or dense temporal captions, \ourmethod adopts a simplified world-state captioning formulation. Under the premise that generative world models should decouple scene constitution from action-driven exploration, captions in this pipeline exclusively summarize the static visual constitution of the environment (\eg, atmospheric conditions, spatial geometry, and layout entities). All temporal progression, dynamic steering, and viewport navigation are delegated entirely to the interactive control tokens.

\begin{table}[!h]
\centering
\vspace{-0.6em}
  \caption{Navigation data composition distribution.}
    \vspace{-1em}
    \renewcommand{\arraystretch}{1.3}
    \footnotesize
    \resizebox{0.48\textwidth}{!}{%
   \begin{tabular}{cccccc}
     \hlineB{2.5}
    \rowcolor{CadetBlue!20} 
    \textbf{DL3DV} & \textbf{SpatialViD} & \textbf{ABot} & \textbf{EgoExo} & \textbf{Sekai} & \textbf{RealEstate} \\
    \hline
    29.22\% & 27.5\% & 25.24\% & 8.76\% & 5.45\% & 3.83\%  \\
     \hlineB{2.5}
   \end{tabular}%
}
  \label{app_fig:data_distribution}
  \vspace{-1.6em}
\end{table} 

\vspace{-1em}
\paragraph{Data Distribution.} The distribution of finally adopted navigation data sources is illustrated in \Cref{app_fig:data_distribution}.

\subsection{Conditioning Schema}

To facilitate versatile interactive control while accommodating practical annotation constraints, \ourmethod adopts a two-stage training curriculum and a mixed-control schema across viewing regimes.

\begin{figure}[!h]
\centering
\vspace{-0.6em}
\includegraphics[width=0.9\linewidth]{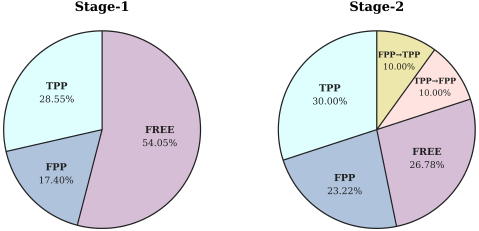}
\vspace{-0.8em}
\caption{View distribution across two stages.}
\label{app_fig:data_schema}
\vspace{-1.8em}
\end{figure}

\paragraph{Two-Stage Training Curriculum.}
As outlined in \Cref{sec:training}, training proceeds in two sequential phases (\Cref{app_fig:data_schema}):
\begin{itemize}[leftmargin=1.4em]
    \item[\ding{192}] \textit{Stage-1 (General Navigation):} \ourmethod is trained on $100$K single-regime navigation clips to learn fundamental physical dynamics and interactive responsiveness across \textsc{Free}, \textsc{Fpp}, and \textsc{Tpp} modes.
    \item[\ding{193}] \textit{Stage-2 (Switching Consolidation):} Rather than fine-tuning solely on switching data, \ourmethod constructs an $8$K curriculum by mixing regular navigation clips with the $500$ switching sequences, where switching clips are oversampled to reinforce transition stability while preserving base navigation capabilities.
\end{itemize}

\begin{figure}[!h]
\centering
\vspace{-0.6em}
\includegraphics[width=\linewidth]{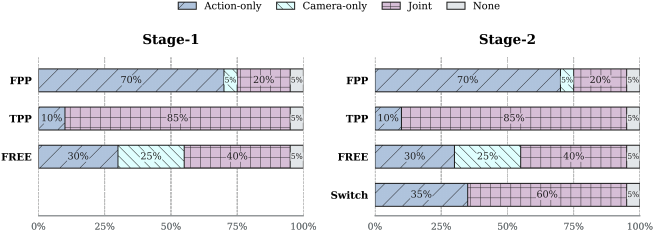}
\vspace{-2em}
\caption{Action/camera distribution across two stages.}
\label{app_fig:data_schema_2}
\vspace{-1.8em}
\end{figure}

\paragraph{Mixed Action-Camera Conditioning.}
In \Cref{sec:action_camera}, ``joint training'' refers precisely to the composite conditioning mixture illustrated in \Cref{app_fig:data_schema_2}, rather than strictly enforcing concurrent dual signals on every training instance. In real-world data, simultaneously obtaining robust annotations for both modalities is often impractical; for instance, loose camera-following in \textsc{Tpp} sequences complicates clean action discretization, and synthetic scenes can introduce camera pose estimation artifacts. 
Therefore, \ourmethod trains across a composite conditioning mixture of \textit{action-only}, \textit{camera-only}, \textit{joint}, and \textit{unconditioned} samples.

\vspace{-1em}
\paragraph{Distribution Transparency.}
The mixture schedules in \Cref{app_fig:data_schema} and \Cref{app_fig:data_schema_2} are reported for empirical transparency \textbf{rather than claimed as an optimal configuration}, reflecting practical trade-offs against data availability. Crucially, all comparative baselines (\eg, additive injection and AdaLN modulation in \Cref{sec:experiments}) follow this identical conditioning schema to guarantee strictly controlled architectural comparisons.

\subsection{More Details of IT-RoPE}
\label{app_sec:it_rope_details}

In this section, we further detail additional mathematical formulations, implementation properties, and efficient computation of Interaction-Transport RoPE (IT-RoPE).

\vspace{-1em}
\paragraph{Transport Accumulation and Prefix Formulation.}
IT-RoPE maintains dual, decoupled transport paths for character action and camera motion ($b \in \{a, c\}$), aligned with the latent temporal grid. Within each continuous valid segment anchored at initial time $s$, the orthogonal step transformations $U_t^b$ accumulate into prefix operators:
\begin{equation}
    P_s^b = I, \qquad P_t^b = P_{t-1}^b U_t^b, \qquad T_{i \rightarrow j}^b = (P_i^b)^\top P_j^b,
\end{equation}
which recovers the ordered interval transport $T_{i \rightarrow j}^b = \prod_{k=i+1}^j U_k^b$ ($i < j$) defined in Eq.~\eqref{eq:interval_transport} and supports both forward and backward temporal relations. Because the step transformations need not commute, the accumulated transport retains order-dependent information about the execution sequence rather than reflecting only endpoint states.

\vspace{-1em}
\paragraph{Gating, Centering, and Initialization.}
The two transport corrections act additively on the base 3D-RoPE attention scores. Specifically, gates $g_i^b$ conditioned on view mode and token type dynamically adjust the relative influence of each control branch, while interval masks $m_{ij}^b \in \{0, 1\}$ restrict corrections to token pairs within contiguous valid execution segments. Centering the transport operator by subtracting the identity matrix $I$ ensures that the correction vanishes whenever $T_{t_i \rightarrow t_j}^b = I$. This holds for same-timestep pairs ($t_i = t_j$) via the prefix formulation, as well as for static intervals where the control parameterization yields identity step transformations ($U_t^b = I$). Furthermore, zero-initializing the branch scale factors $\lambda^b$ preserves the base attention scores at initialization.

\vspace{-1em}
\paragraph{Factorized Attention Computation.}
By exploiting the orthogonality of the prefix operators $P_t^b$, the centered inner product expands into:
\begin{equation}
    (q_i^b)^\top (T_{t_i \rightarrow t_j}^b - I) k_j^b = (P_{t_i}^b q_i^b)^\top (P_{t_j}^b k_j^b) - (q_i^b)^\top k_j^b.
\end{equation}
Within each continuous valid segment, the correction decomposes into the difference between prefix-transformed and untransformed query-key dot products. This factorization avoids explicitly constructing pairwise transformation matrices for all frame intervals, allowing the prefix accumulation and corresponding query-key transforms to scale linearly with the temporal length and token count.

\section{Experimental Details}
\label{app_sec:experiemntal_details}


\subsection{Evaluation Benchmark}
\label{app_sec:evaluation_bench}

We evaluate \ourmethod on WBench \cite{ying2026wbench}, a benchmark for interactive video world models comprising $289$ cases and $1,058$ interaction turns. Each case provides an initial image, a world description, and multi-turn control sequences. We use its \textsc{Navigation} split and \textsc{Fpp}$\leftrightarrow$\textsc{Tpp} perspective-switching subset.

\vspace{-1em}
\paragraph{Navigation Suite.}
WBench benchmarks trajectory-following navigation across five core dimensions:
\begin{itemize}[leftmargin=*]
    \item \textit{Quality}: Assesses perceptual aesthetics, visual fidelity, motion dynamics, temporal flickering, and inter-frame motion smoothness.
    \item \textit{Setting}: Measures contextual adherence to the specified prompt, including scene appearance preservation, environmental style consistency, and subject persistence.
    \item \textit{Interaction}: Evaluates motion responsiveness and command fidelity by comparing estimated camera trajectories with action-derived reference trajectories under translational, rotational, and symmetric steering commands.
    \item \textit{Consistency}: Quantifies structural and geometric coherence, capturing character identity stability, absence of shot cuts, and photometric/geometric consistency across viewpoint transitions.
    \item \textit{Physical}: Assesses adherence to real-world physics, commonsense dynamics, and causal coherence throughout interactive rollouts.
\end{itemize}
In addition to individual dimension scores, the overall \textsc{Navigation} average is reported.

\vspace{-1em}
\paragraph{Perspective Switching.}
For each requested \textsc{Fpp}$\leftrightarrow$\textsc{Tpp} transition, WBench evaluates the visual rollout to determine whether viewpoint transition smoothness, terminal perspective, and framing geometry satisfy the target specifications. Binary success decisions are aggregated across all evaluation turns and reported as an independent metric to isolate dynamic viewpoint adaptability from trajectory tracking.

All evaluations follow the official WBench protocol, employing its standardized scoring functions, decision thresholds, and aggregation rules without modification.

\subsection{Setup of Analysis in \Cref{sec:method}}
\label{app_sec:method_setup}

This section details the experimental configurations, evaluation datasets, and diagnostic metrics employed in the analyses of \Cref{sec:method}.

\vspace{-1em}
\paragraph{View-Dependent Motion Dynamics (\Cref{fig:42_a}).}
To isolate regime-dependent motion characteristics, $100$ first-person navigation scenarios are randomly sampled from WBench. For each scene, the world-state caption is held static, and the action command sequence is fixed uniformly to continuous forward movement (\texttt{W}). The model is then prompted under contrasting view modes (\textsc{Free} \textit{vs.} \textsc{Fpp}) to observe the resulting physical trajectories. 

Camera trajectories $T_t = [R_t, p_t; 0, 1]$ are reconstructed using ViPE \cite{huang2025vipe}. Positions are expressed relative to the initial frame coordinates $q_t = R_0^\top(p_t - p_0)$, with $u_t = (q_{t,x}, -q_{t,y})$ representing transverse (lateral and vertical) coordinates. The smooth trajectory trend is isolated via a second-order Savitzky-Golay filter with a 33-frame window, yielding the transverse residual:
\begin{equation}
    r_t = u_t - \mathrm{SG}_{33,2}(u)_t.
\end{equation}
The oscillation amplitude is quantified via normalized Root Mean Square error ($\text{RMS}_{\text{osc}}$):
\begin{equation}
    \text{RMS}_{\text{osc}} = \frac{100}{L} \sqrt{\frac{1}{|\mathcal{I}|} \sum_{t \in \mathcal{I}} \left( r_{t,x}^2 + r_{t,y}^2 \right)},
\end{equation}
where $L = |q_{T-1,z}|$ denotes the net forward displacement, and $\mathcal{I}$ is the interior interval excluding $0.5$ seconds at sequence boundaries. This metric captures transverse body-sway and vertical bobbing while remaining invariant to global forward drift.

\vspace{-1em}
\paragraph{Emergent View Switching (\Cref{fig:42_b}).}
The quantitative validation evaluates viewpoint transition fidelity on the official WBench \textsc{Fpp}$\leftrightarrow$\textsc{Tpp} split, adhering strictly to the VLM-based transition criteria and aggregation protocol described in \Cref{app_sec:evaluation_bench}.

\vspace{-1em}
\paragraph{Action \textit{vs.} Camera Control Disentanglement (\Cref{fig:43}).}
The comparative analysis of action and camera control is conducted on the WBench \textsc{Navigation} suite, with third-person (\textsc{Tpp}) instances excluded to isolate egocentric/free dynamics. Evaluation spans four metrics: two local command-fidelity diagnostics and two global consistency measures.
\begin{itemize}[leftmargin=*]
    \item \textit{Translation Accuracy}: Turn sequences are segmented into non-overlapping $0.5$-second windows $w=[a, b]$. Camera poses are tracked via MegaSaM \citep{li2025megasam}. For each window, the generated local displacement $v_w = R_a^\top(p_b - p_a)$ is compared against the action-integrated reference $v_w^* = (R_a^*)^\top(p_b^* - p_a^*)$. A window is classified as accurate if the directional deviation satisfies $\arccos\left(\frac{v_w^\top v_w^*}{\|v_w\|_2 \|v_w^*\|_2}\right) \le 30^\circ$ and the displacement exceeds a minimum motion threshold $\|v_w\|_2 \ge 0.004 d_0 \Delta t_w$, where $d_0$ is the median initial positive depth and $\Delta t_w$ is the window duration.
    \item \textit{Rotation Accuracy}: Using identical temporal windows, relative rotation vectors $\omega_w = \log(R_a^\top R_b)^\vee$ and references $\omega_w^* = \log((R_a^*)^\top R_b^*)^\vee$ are extracted in $\mathfrak{so}(3)$. A window is correct if the angular orientation deviation is within $30^\circ$ and the rotational magnitude satisfies $\|\omega_w\|_2 \ge (\pi / 180) \Delta t_w$. Accuracy scores are averaged over valid active windows and scaled to a percentage.
    \item \textit{Geometric Consistency}: Following WBench, Depth Anything 3 \citep{lin2025depth} provides predicted depths $D_i$, intrinsics $K_i$, and poses $T_i$. Source pixels are back-projected and reprojected into target frame $j$: $X_j = T_j^{-1} T_i (D_i(u) K_i^{-1} u)$, measuring relative depth disparity $e_j = \mathrm{median}[|z' - D_j(u')| / D_j(u')]$. The consistency score is computed as $\text{GC} = 100 / (1 + \mathrm{mean}_j(e_j))$.
    \item \textit{Photometric Consistency}: Following WBench, colors from source frames are splatted into target frame $j$ via depth ordering. Denoting $\hat{I}_j$ as the splatted image and $\Omega_j$ as valid pixels, the average $\overline{\text{PSNR}}$ is evaluated across frames with $|\Omega_j| \ge 100$. Photometric consistency is aggregated as $\text{PC} = 100 \cdot (1 - 10^{-\overline{\text{PSNR}}/20})$.
\end{itemize}

\subsection{Setup of Analysis in \Cref{sec:experiments}}
\label{app_sec:experiment_setup}

Additive injection and AdaLN modulation are re-implemented following the conditioning strategies of ABot-World \cite{jiang2026abot} and LingBot-World \cite{gao2026infinite}, respectively. For explicit view conditioning in these baselines, the categorical view mode is embedded at each latent timestep, spatially broadcast, and fused with the available action and camera representations. The resulting composite features are added directly to the visual patch tokens for additive injection, or projected to predict affine modulation parameters for AdaLN. In \S\ref{sec:5.3} and \S\ref{sec:5.4}, the default \ourmethod is evaluated prior to stage-2 switching consolidation to demonstrate the core architectural advantages of in-context conditioning and transport modeling.


\end{document}